\documentclass[times,twocolumn,final]{elsarticle}

\usepackage{medima}
\usepackage{framed,multirow}
\usepackage{amsmath}
\usepackage{array}
\usepackage{arydshln}
\usepackage{float}
\usepackage{amssymb}
\usepackage{latexsym}
\usepackage{comment}
\usepackage{xcolor}
\usepackage{graphicx}
\usepackage{babel}
\usepackage{subfig}
\usepackage{pgfplots}
\usepgfplotslibrary{statistics}
\usepackage{dblfloatfix}
\usepackage{hyperref}

\definecolor{newcolor}{rgb}{.8,.349,.1}

\newcommand{\tikzcircle}[2][red,fill=red]{\tikz[baseline=-0.65ex]\draw[#1,radius=#2] (0,0) circle ;}%

\journal{Medical Image Analysis}

\begin{document}
\definecolor{bblue}{HTML}{4F81BD}
\definecolor{rred}{HTML}{C0504D}
\definecolor{ggreen}{HTML}{9BBB59}
\definecolor{ppurple}{HTML}{9F4C7C}
\definecolor{bittersweet}{HTML}{00B3B8}

\verso{de Boisredon \textit{et~al.}}

\begin{frontmatter}

\title{Image-level supervision and self-training for transformer-based cross-modality tumor segmentation}%

\author[1]{Malo \snm{Alefsen de Boisredon d'Assier}}
\author[3]{Eugene \snm{Vorontsov}}
\author[1,2]{William Trung \snm{Le}}
\author[1,2]{Samuel \snm{Kadoury}}

\address[1]{Polytechnique Montreal, Montreal, QC, Canada}
\address[3]{Paige, Montreal, QC, Canada}
\address[2]{Centre de Recherche du Centre Hospitalier de l’Université de Montréal, Montreal, QC, Canada}

\received{}
\finalform{}
\accepted{}
\availableonline{}
\communicated{}

\begin{abstract}
Deep neural networks are commonly used for automated medical image segmentation, but models will frequently struggle to generalize well across different imaging modalities. This issue is particularly problematic due to the limited availability of annotated data, making it difficult to deploy these models on a larger scale. To overcome these challenges, we propose a new semi-supervised training strategy called MoDATTS. Our approach is designed for accurate cross-modality 3D tumor segmentation on unpaired bi-modal datasets. An image-to-image translation strategy between imaging modalities is used to produce annotated pseudo-target volumes and improve generalization to the unannotated target modality. We also use powerful vision transformer architectures and introduce an iterative self-training procedure to further close the domain gap between modalities. MoDATTS additionally allows the possibility to extend the training to unannotated target data by exploiting image-level labels with an unsupervised objective that encourages the model to perform 3D diseased-to-healthy translation by disentangling tumors from the background. The proposed model achieves superior performance compared to other methods from participating teams in the CrossMoDA 2022 challenge, as evidenced by its reported top Dice score of $0.87\pm0.04$ for the VS segmentation. MoDATTS also yields consistent improvements in Dice scores over baselines on a cross-modality brain tumor segmentation task composed of four different contrasts from the BraTS 2020 challenge dataset, where $95\%$ of a target supervised model performance is reached. We report that 99\% and 100\% of this maximum performance can be attained if 20\% and 50\% of the target data is additionally annotated, which further demonstrates that MoDATTS can be leveraged to reduce the annotation burden.
\end{abstract}


\end{frontmatter}


\section{Introduction}
\label{sec1}

Deep learning has shown outstanding performance and potential in various medical image analysis applications \citep{image_analysis}. Notably, it has been successfully leveraged in medical image segmentation, showing equivalent accuracy to manual expert annotations \citep{image_segmentation}. However, these breakthroughs are tempered by the issue of performance degradation when models face data from an unseen domain \citep{shift}. This problem is particularly important in medical imaging, where distribution shifts are common. Annotating data from all domains would be inefficient and intractable, notably in image segmentation where expert pixel-level labels are expensive and difficult to produce \citep{challenges}. Building models that can generalize well across domains without any additional annotations is thus a challenge that needs to be addressed. Specifically, cross-modality generalization is a key contribution towards the reduction of the data dependency and the usability of deep neural networks at a larger scale. Applications of such models are manifold, as it is common that one imaging modality lacks annotated training examples. For instance, acquisition of contrast-enhanced T1-weighted (T1ce) MR images is the most commonly used protocol for Vestibular Schwannoma (VS) detection. Accurate diagnosis and delineation of VS is of considerable importance to avoid boundless tumor growth, which can lead to irreversible hearing loss. However, in order to reduce scan time in T1ce imaging and alleviate the risks associated with the use of Gadolinium contrast agent, high resolution T2-weighted (hrT2) has recently gained popularity in clinical workflows \citep{VS_acq}. Existing annotated T1ce databases can thus be leveraged to address the lack of training data for VS segmentation on hrT2 images. Furthermore, such models could be used for anomaly detection across modalities and pathologies, if the associated lesions show similar patterns. An example is to use pixel-level annotations of brain gliomas in MRIs to learn the distribution of intraparenchymal hemorrhages on CT scans \citep{mr_ct}.

\begin{figure*}[!t]
\centering
\includegraphics[scale=0.272]{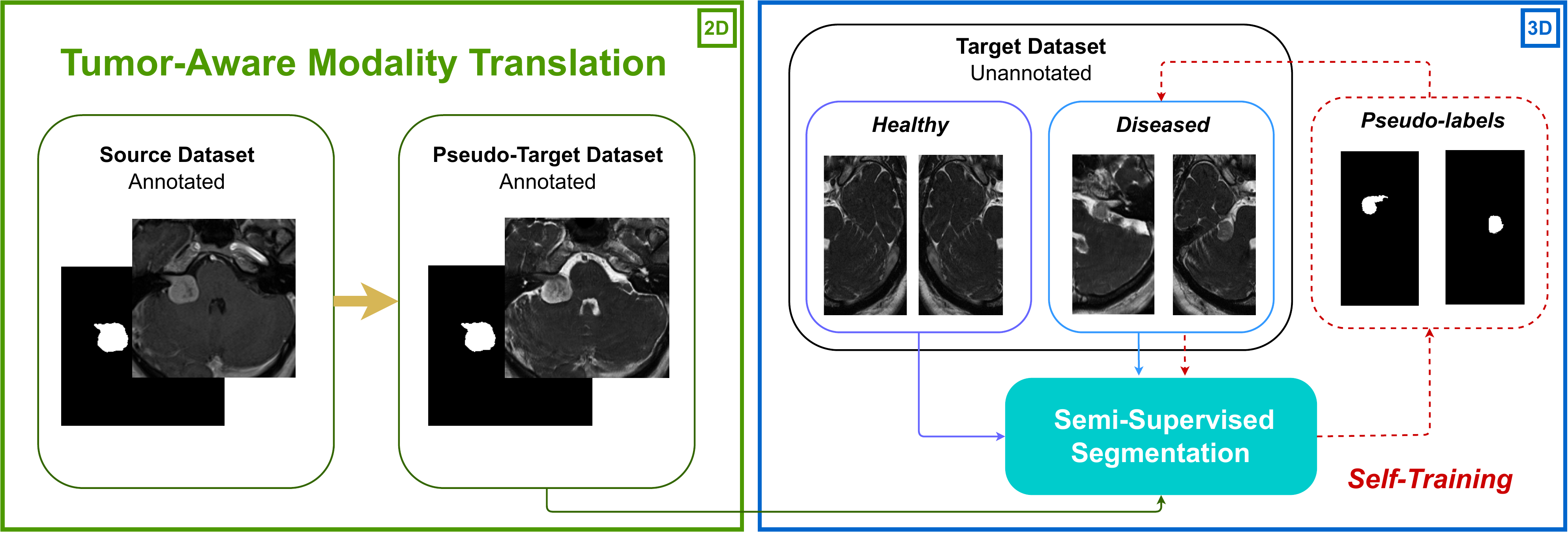}
\caption{Overview of MoDATTS. Stage 1 (green) consists in generating realistic target images from source data by training cyclic cross-modality translation.  In stage 2 (blue), segmentation is trained in a semi-supervised approach on a combination of synthetic and original target modality images. Finally, pseudo-labeling is performed and the segmentation model is refined through several self-training iterations.}
\label{Mgenseg}
\end{figure*}

Recently, a method to segment images from a wide range of contrasts without any retraining or fine-tuning was proposed by \cite{synthseg}. Using a generative approach conditioned on segmentations they synthetically generate images of random contrasts and resolutions, used at a later stage to train a segmentation network robust to highly heterogeneous data. However, although this domain randomisation strategy demonstrates improved generalization capability for brain parcellation, the model performance when exposed to tumours and pathologies was not quantified. More commonly, the key challenge of cross-modality generalization can be tackled through unsupervised domain adaptative (UDA) methods, which aim at leveraging the information learned from a ``source'' domain with abundant labeled data to improve the performance of a model on a ``target'' domain where labeled data is scarce or unavailable \citep{dom_adapt}. In medical imaging applications, several UDA strategies are based on feature space alignment and have been widely adopted for cross-modality organ segmentation (e.g. delineation of cardiac structures \citep{card_1,card_2,card_3}). More widespread UDA models are based on generative strategies and tackle the issue by teaching the model to perform image-to-image translations \citep{image-to-image} between modalities. Through adversarial training, annotated synthetic pseudo-target images can be generated from annotated source modality images and used to train a segmentation network. These methods demonstrate satisfactory results but solely rely on pixel-level annotations for source modality images. Furthermore, due to dataset and training resource constraints, these end-to-end models tend to be limited to 2D. While modality translation can be performed in 2D without performance loss, segmentation tasks highly benefit from computations on 3D volumes rather than 2D slices.

\begin{figure*}[!t]
\centering
\includegraphics[scale=0.183]{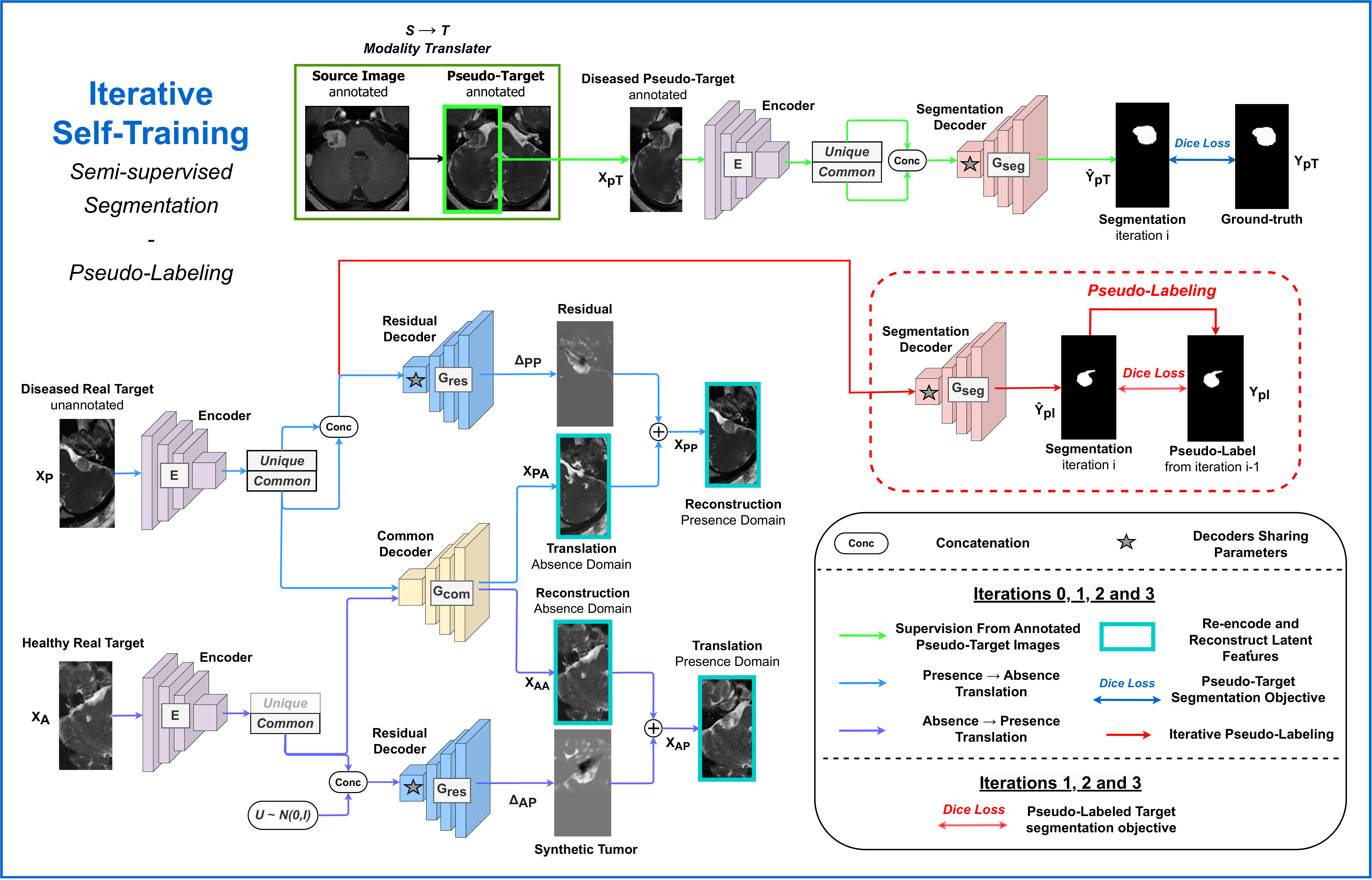}
\caption{Overview of the segmentation stage. A segmentation decoder $G_{seg}$ is trained for tumor segmentation on annotated pseudo-target images resulting from stage 1. Simultaneously a common decoder $G_{com}$ and a residual decoder $G_{res}$ are jointly trained for unsupervised tumor delineation on real target images by performing diseased $\rightarrow$ healthy and healthy $\rightarrow$ diseased translations. The segmentation decoder shares parameters with the residual decoder to benefit from this unsupervised objective. Finally, The segmentation encoder-decoder $G_{seg}\circ E$ is used to generate pseudo-labels for unannotated target images. The model is then further refined through several self-training iterations.}
\label{self_training}
\end{figure*}

Hence we propose in this paper MoDATTS, a new \textbf{M}odality \textbf{Do}main \textbf{A}daptation \textbf{T}ransformer-based pipeline for \textbf{T}umor \textbf{S}egmentation which aims at bridging the gap between a partially annotated source modality and an unannotated target modality. As illustrated in Fig. \ref{Mgenseg}, our model comprises two stages for training. In the first stage a 2D network is taught to translate between imaging modalities, to eventually generate pseudo-target images from the source brain volumes. The translation generators are bounded to preserve the tumor information during the modality transfer by sharing the latent representations with segmentation decoders (see Fig. \ref{tum_aware_mod_trans}). The resultant annotated synthesized target images are then used in a second stage to teach a 3D network to perform the segmentation task (see Fig. \ref{self_training}). To alleviate the need for source annotations and extend the training to original target images, we incorporate an unsupervised anomaly detection objective on the target modality. This is done by leveraging  a 2D generative strategy (GenSeg) that uses image-level ``diseased'' or ``healthy'' labels for semi-supervised segmentation \citep{genseg}. Similarly to low-rank atlas based methods \citep{rank1,rank2,rank3} the model is taught to find and remove the lesions, which acts as a guide for the segmentation. An iterative self-training procedure is also implemented to further close the gap between source and target modalities. Finally, MoDATTS leverages powerful vision transformer architectures to enhance the modality translation and segmentation.
\\\\\
Considering the challenge of cross-modality tumor segmentation, our main contributions are as follow :
\begin{enumerate}
    \item We propose a tumor-aware modality translation training procedure that can accurately retain the shape of lesions.
    \item We develop a 3D segmentation network that can leverage volumes known to be healthy, and explore its potential for unsupervised tumor delineation on cross-modality segmentation tasks.
    \item We build our domain adaptation framework with effective vision transformer architectures.
    \item The proposed model has the ability to augment the training set using pseudo-labeling and self-training mechanisms.
\end{enumerate}

MoDATTS is evaluated on two distinct cross-modality tumor segmentation tasks: (i) a customized version of the BraTS 2020 dataset \citep{b3,b2,b1}, where each of the four contrast sequences (T1, T2, T1ce, and FLAIR) were treated as separate modalities, and (ii) the CrossMoDA 2022 data challenge \citep{crossmoda_1,crossmoda_2}. We demonstrate that our model can better generalize than other state-of-the-art methods to the target modality and yield robust performance even with few source modality annotations.

\section{Related Works}

\subsection{Unsupervised domain adaptation}

Domain adaptation has emerged as a popular solution to address the common issue of domain shifts and heterogeneity in medical imaging. By minimizing distribution differences between related but different domains, UDA methods facilitate the use of machine learning models across varied medical image datasets \citep{Domain}. Latent space alignment strategies have been widely adopted in different applications to deal with heterogeneity between sets of images from different centers (e.g. knee tissue segmentation \citep{knee} or mass detection on mammograms \citep{mass}) or with different modalities (e.g. cardiac structures segmentation on MRI using CT scans \citep{card_3}). In these models, an encoder is trained to learn modality invariant representations of the images either through divergence minimization of the feature distributions or adversarial training on the latent spaces. A segmentation decoder trained on annotated source data is then bounded to produce consistent segmentation maps for the target images. For tumor segmentation tasks, generative approaches based on cross-modality translation are more frequent and will be reviewed in next section. 

To alleviate the need for source data availability during the adaptation stage, some source-free domain adaptation methods have been developed. Using a source segmentation model, \cite{source_free_1} proposed to transform target images into high-quality source-like images with batch norm constraints. As low-frequency components in the Fourier domain can represent style information, refinement of the generated images is achieved with the mutual Fourier Transform. Feature-level and output-level alignment is then performed based on the generated paired source-like and target images. \cite{source_free_2} used a pre-trained tumor segmentation model on source T2-weighted MRI brain images and fine-tuned its parameters on the target domain (T1, T1-weighted or FLAIR) by explicitly enforcing high order batch statistics consistency and minimizing the self-entropy of predictions on the target distribution. The adaptation phase proposed by \cite{source_free_3} involved minimizing a loss function that incorporates the Shannon entropy of predictions and a prior based on the class-ratio in the target domain. However, these approaches under-perform in comparison to state-of-the-art generative methods and often relies on image-level labels incurring substantial annotation costs \citep{source_free_3}.

\subsection{Style transfer and cross-modality segmentation}

Style transfer neural networks, which involve transferring the visual appearance (or style) of one image to another while preserving the content of the latter, were first introduced by \cite{style_1}. Their approach enables the generation of novel images with high perceptual quality that combine the content of any given photograph with the visual style of various well-known artworks. Such models can be leveraged for domain adaptation purposes in medical image applications by generating synthetic images in the target domain to supervise a target modality segmentation model. Notably, the CycleGAN model proposed by \cite{cyclegan} became the standard for transfers between imaging modalities. CycleGAN is an unpaired bidirectional image translation network based on generative adversarial training, and preserves content specific information through cyclic pixel-level reconstruction constraints. Several works proposed a domain adaptation framework based on a CycleGAN-like approach to perform modality translation \citep{x_ray,SynSeg,Synergistic,self,attent,constrained}. Segmentation is jointly trained in an end-to-end manner on the labeled synthetic target images translated from the annotated source modality. Alternatively modality translation can be combined with latent space alignment to further regularize the model. \cite{disentangle} retained the principle of cyclic modality translations but proposed to jointly disentangle the domain specific and domain invariant features between each modality and train a segmenter on top of the domain invariant features. Similarly, \cite{prior_matching} proposed a VAE-based feature prior matching mechanism to learn domain invariant features while training for modality translation and segmentation.
\\\\
Note that in these methods the modality translation networks are able to maintain the structures of interest (e.g. the tumours) by integrating features from the segmentation network. Due to memory constraints, performing segmentation end-to-end with modality translation on full 3D volumes is not tractable. Thus, performing domain adaptation in a two-stage manner with 2D tumor-aware modality translation followed by 3D segmentation is an adequate setting.

\begin{figure*}[!t]
\centering
\includegraphics[scale=0.262]{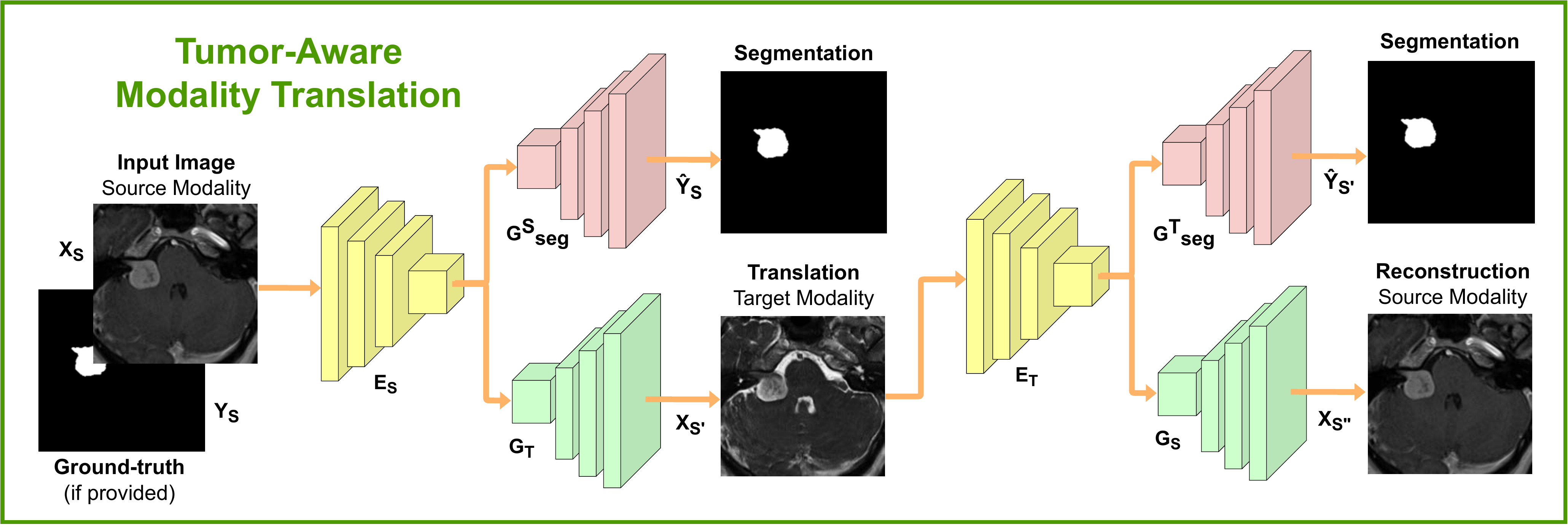}
\caption{Overview of the proposed tumor-aware cross-modality translation. The model is trained to translate between modalities in a CycleGAN approach. $G_T\circ E_S$ and $G_S\circ E_T$ encoder-decoders respectively perform S $\rightarrow$ T and T $\rightarrow$ S modality translations. Same latent representations are shared with co-trained segmentation decoders $G_{seg}^{S}$ and $G_{seg}^{T}$ to preserve the semantic information related to the tumors. Note that the T $\rightarrow$ S $\rightarrow$ T translation loop is not represented for readability. We precise that the latter does not yield segmentation predictions since we assume no annotations are provided for the target modality.}
\label{tum_aware_mod_trans}
\end{figure*}

\subsection{Self-training}

Self-training is a semi-supervised learning technique with pseudo-labeling. A teacher model trained on labeled data is used to produce pseudo-labels on unlabeled data. Only pseudo-labels that the model predicts with a high probability are retained. This process can be iterated on the expanded label set, generating additional pseudo-labels. Augmenting the training set with pseudo-labeled examples increases the model's robustness towards out-of-distribution data \citep{self_training}. It was shown to have great potential in leveraging unlabeled data in semantic segmentation applications \citep{self_seg_1,self_seg_2}. It is also a suitable candidate for improvements in domain adaptation tasks \citep{self_adapt_1,self_adapt_2,self_adapt_3,self_adapt_4}. Self-training was introduced for cross-modality segmentation in the context of the CrossMoDA 2021 domain adaptation challenge \citep{cm_1}. In the reiteration of the challenge in 2022, self-training was a core strategy among the top ranked methods \citep{MAI,LATIM}.

\section{Methods}

Let us consider the scenario where we have a set of images ${X_T}$ without pixel-level tumor annotations for a ``target'' modality T. The objective of this work is to learn consistent segmentations on the target data using a second set of images ${X_S}$ of the ``source'' modality S, that is partially or totally annotated with labels ${Y_S}$. Note that the datasets are considered to be unpaired.

\subsection{Tumor-aware cross-modality translation}
\label{stage_1}
The first phase of our model consists in augmenting the source images into realistic pseudo-target images, so that the pixel-level annotations available in the source modality can be reused to train a segmentation network on the target modality. Based on the CycleGan model \citep{cyclegan}, we perform modality translations via two distinct encoder-decoder networks (see Fig. \ref{tum_aware_mod_trans}). Encoders $E_S$ and $E_T$ are used to encode source and target modality images, respectively. Combined with $E_S$, a decoder $G_T$ enables performing S$\rightarrow$T modality translation, while $E_T$ and a second decoder $G_S$ performs the T$\rightarrow$S modality translation. To preserve the anatomical contents, the model is forced to reconstruct the original images after mapping back to the original modality. This is referred to as cycle-consistency. We note $X_{S'}=G_T\circ E_S(X_{S})$ and $X_{S''}=G_S\circ E_T(X_{S'})$, respectively the translation and reconstruction of $X_{S}$ in the S$\rightarrow$T$\rightarrow$S translation loop. Similarly we have $X_{T'}=G_S\circ E_T(X_{T})$ and $X_{T''}=G_T\circ E_S(X_{T'})$ for the T$\rightarrow$S$\rightarrow$T cycle. We specify that $\circ$ is the composition operation.\vspace{-0.2cm}
\\\\
Because image reconstruction from cycle-consistency is imperfect in practice \citep{distrib}, the model is guided to specifically retain detailed geometrical tumor structures by incorporating two segmentation decoders $G_{seg}^S$ and $G_{seg}^T$. This has shown to be efficient for two-stage domain adaptation methods \citep{cosmos}. The segmentation decoders share the same latent input representation as the modality decoders, which constrains the encoders to learn features that encompass the tumors information. For each annotated source image, the model outputs a segmentation map of the original image ($\hat{Y}_S=G_{seg}^S\circ E_S(X_{S})$) and the image's translation to the target domain ($\hat{Y}_{S'}=G_{seg}^T\circ E_T(X_{S'})$).
\\\\
The loss function for this stage is therefore composed of` three terms :
\begin{enumerate}
    \item An \textbf{adversarial objective} based on the hinge loss that aims at discriminating between real and generated images of the same modality: 
    \begin{equation}
    \label{hinge}
        \begin{split}
        \hspace{-0.8cm}\mathcal{L}_{adv}^{mod} = &\sum_{m\in\{S, T\}}  \min_{G}\max_{D}\mathbb{E}_{X_m}\left(\min\left(0,D_m(X_m)-1\right)\right)\\
        &-\mathbb{E}_{\hat{X}_m}\left(\min\left(0,-D_m(\hat{X}_m)-1\right)\right)-\mathbb{E}_{\hat{X}_m}D_m(\hat{X}_m)
        \end{split}
    \end{equation}
    \item A \textbf{reconstruction loss} enforcing cycle consistency:
    \begin{equation}
    \mathcal{L}_{cyc}=\left\Vert X_{S}-X_{S''} \right\Vert_1 + \left\Vert X_{T}-X_{T''} \right\Vert_1
    \end{equation}
    \item A \textbf{segmentation objective}, based on a differentiable soft Dice loss like in \cite{soft_dice}:
    \begin{equation}    \mathcal{L}_{seg}^{mod}=Dice\left({Y_S},{\hat{Y}_S}\right)+Dice\left({Y_S},{\hat{Y}_{S'}}\right)
\end{equation}
\end{enumerate}

The overall translation loss $\mathcal{L}_{trans}$ is a weighted sum of the aforementioned terms: 
\begin{equation} 
\mathcal{L}_{trans} = \lambda_{seg}^{mod}\mathcal{L}_{seg}^{mod}+ \lambda_{adv}^{mod}\mathcal{L}_{adv}^{mod}+\lambda_{cyc}\mathcal{L}_{cyc}
\end{equation}
Note that to facilitate the hyper-parameter search, weights are normalized so that their sum always equals to 1.

\subsection{Target modality segmentation}

\paragraph{\textbf{Supervision from pseudo-target data}}

Once modality translation is learned, we yield a dataset of pseudo-target images $X_{pT}$ and their corresponding pixel-level annotations $Y_{pT}$. Like in state-of-the-art methods, prior to self-training iterations we train a 3D segmentation network by teaching an encoder $E$ and a decoder $G_{seg}$ the segmentation task on this synthetic data. Based on images $X_{pT}$, we predict segmentation maps $\hat{Y}_{pT}=G_{seg}\circ E(X_{pT})$ that can be compared to the ground-truths $Y_{pT}$ for model optimization. The corresponding loss function is termed :
    \begin{equation}    \mathcal{L}_{seg}^{pT}=Dice\left({Y_{pT}},{\hat{Y}_{pT}}\right)
    \label{eq:synthetic_supervision}
\end{equation}

\paragraph{\textbf{Semi-supervised segmentation}}

Unlike prior methods that simply train the segmentation network on the annotated pseudo-target images before performing self-training, we propose a semi-supervised approach. By using the GenSeg training strategy \citep{genseg}, we believe the model can better fit the target modality distribution than with only supervision from the pseudo-target data which may still suffer from a small distribution shift. This also allows us to model relevant tumor representations even when only few source images have pixel-level annotations.

To localize lesions, we use image-level labels that describe whether an image contains a lesion or not. These ``diseased'' and ``healthy'' labels can be efficiently leveraged by a generative model by translating between presence and absence (of tumor lesions) domains, referred to as P and A. In this setup, we seek to separate the information that is shared between the two domains (A and P) from the information that is specific to domain P (that is, separate out the lesions). We therefore divide the latent representation of each image into two distinct codes: ${c}$ and ${u}$. The common code ${c}$ contains information that is inherent to both domains, such as organs and other structures, while the unique code ${u}$ stores features specific to domain P, such as tumor shapes and localization.

\paragraph{\textbf{Presence to absence translation}}

Given an original image of the target modality in the presence domain $X_{P}$, we use the encoder $E$ to compute its latent representation ${[}{c_{P}},{u_{P}}{]}$. A common decoder $G_{com}$ interprets the common code ${c_{P}}$ and generates a healthy version ${X_{PA}}$ of that image by removing the apparent tumor region. At the same time, a residual decoder $G_{res}$ employs both common and unique codes to produce a residual image ${\Delta_{PP}}$, which represents the additive modification required to shift the generated healthy image back to the presence domain. In other words, the residual is the disentangled tumor that can be added to the generated healthy image to create a reconstruction ${X_{PP}}$ of the initial diseased image:
\begin{equation}
X_{PA}=G_{com}({c_{P}}),
\end{equation}
\vspace{-0.4cm}
\begin{equation}
\Delta_{PP}=G_{res}({c_{P}},{u_{P}}),
\end{equation}
\vspace{-0.3cm}
\begin{equation}
X_{PP}=X_{PA}+{\Delta_{PP}}.
\end{equation}

\paragraph{\textbf{Absence to presence translation}}

In parallel, a similar process is implemented for images in the healthy domain. Given an original target image $X_{A}$ in the absence domain A, a translated version in domain P is generated. Hence, a synthetic tumor ${\Delta_{AP}}$ is created by sampling a code from a prior distribution $\mathcal{N}(0,{I})$ and substituting the encoded unique code for that image. The reconstruction $X_{AA}$ of the original image in domain A and the synthetic diseased image $X_{AP}$ in domain P are calculated from the encoded features ${[}{c_{A}},{u_{A}}{]}$ in the following manner:
\begin{equation}
X_{AA}=G_{com}({c_{A}}),
\end{equation}
\begin{equation}
X_{AP}=X_{AA}+G_{res}({c_{A}},{u}\sim\mathcal{N}(0,{I})).
\end{equation}
A tumor can have various appearances, which means that translating from the absence to the presence domain requires a one-to-many mapping. For this reason, the unique code is replaced by a code sampled from a normal distribution $\mathcal{N}(0,{I})$. Each different sampled unique code can then be interpreted by the residual decoder as a different tumor. Additionally, we reconstruct the latent representations of the generated images in both translation directions to ensure that the information from the original image is preserved. Note that in the absence-to-presence direction this enforces the distribution of unique codes to match the prior $\mathcal{N}(0,{I})$. Indeed, we make $u_{AP}$ match $u$, where $u_{AP}$ is obtained by encoding the fake diseased sample $X_{AP}$ produced with random sample u. It is worth noting that translating from the absence to the presence domain indirectly augments the target modality data, which in turn improves the domain adaptation.

\paragraph{\textbf{Weight sharing}} We use a configuration where the segmentation decoder $G_{seg}$ shares most of its weights with the residual decoder $G_{res}$ and only differs from the latter by a distinct set of normalization parameters and the addition of a classifying layer. Therefore, through the Absence and Presence translations the segmentation decoder is implicitly learning how to disentangle the tumors from the background on original target modality samples. Additional supervision from the pseudo-target data is still required to teach the segmentation decoder how to transform the resulting residual representation into appropriate segmentation maps. However the requirement for source pixel-level annotations is reduced in comparison to usual domain adaptation methods.

\paragraph{\textbf{Loss function}}

To enforce the diseased-healthy translation we rely on the three components exposed below. Note that the synthetic images $X_{pT}$ are excluded from these terms:
\begin{enumerate}
    \item A healthy-diseased translation \textbf{adversarial loss}. We build a hinge loss $\mathcal{L}_{adv}^{gen}$ like in Eq. \ref{hinge} aiming at discriminating between pairs of real/synthetic images of the same output domain i.e.  ${X_A}$ vs ${X_{PA}}$ and ${X_P}$ vs ${X_{AP}}$.
    \item A pixel-level \textbf{image reconstruction loss} $\mathcal{L}_{rec}$ to regularize the translation between A and P domains :
    \begin{equation}
\mathcal{L}_{rec}=\left\Vert X_{AA}-X_{A} \right\Vert_1 + \left\Vert X_{PP}-X_{P} \right\Vert_1
\end{equation}
    \item A \textbf{latent code reconstruction loss} $\mathcal{L}_{lat}$ that forces the model to preserve information, enforcing a one-to-one correspondence between latent codes and their corresponding images.
\end{enumerate}

The image and latent code reconstruction losses together prevent mode collapse and makes sure that when a tumor is added or removed, the background tissue remains the same. To train the 3D segmentation model, we define a global weighted sum $\mathcal{L}_{Seg}^{init}$ that encompass the diseased-healthy translation losses along with the synthetic supervision term $\mathcal{L}_{seg}^{pT}$ (Eq.~\ref{eq:synthetic_supervision}):
\begin{equation} 
\mathcal{L}_{Seg}^{init}=\lambda_{adv}^{gen}\mathcal{L}_{adv}^{gen}+\lambda_{lat}\mathcal{L}_{lat}+\lambda_{rec}\mathcal{L}_{rec}+\lambda_{seg}^{pT}\mathcal{L}_{seg}^{pT}
\end{equation}
In the same way as for modality translation in the first phase (Sec. \ref{stage_1}), weights are normalized so that their is sum equal to 1 in order to ease hyper-parameter tuning.

\paragraph{\textbf{Self-training}}

At this stage the model has already been trained on real target modality images through the diseased-healthy translation objective. However, the segmentation decoder would specifically benefit from tuning on the original data as it was essentially trained on the synthetic pseudo-target images. We thus further include non-annotated original data to the segmentation objective with a self-training procedure as in \cite{cosmos}. To do so, the segmentation model is used to output probability maps for each target domain images. These are then thresholded with a value $\alpha$ to keep only the predictions in which the model has a high confidence. The resulting pseudo-labels $Y_{pl}$ are considered as new ground-truth annotations for the unannotated target images for finetuning the segmentation model. This procedure can be repeated $k$ times to iteratively refine the pseudo-labels and improve the model segmentation performance on the unannotated modality. During the $i^{th}$ self-training iteration we thus compute predictions $\hat{Y}_{pl}$ that can be compared to the pseudo-labels ${Y}_{pl}^{i-1}$ resulting from the $(i-1)^{th}$ training stage. This is done with an additional self-training segmentation term $\mathcal{L}_{seg}^{st}=Dice\left({Y_{pl}^{i-1}},{\hat{Y}_{pl}}\right)$. Hence we obtain the following global loss for self-training iterations:
\begin{equation} 
\mathcal{L}_{Seg}^{ST}=\mathcal{L}_{Seg}^{init}+\lambda_{seg}^{st}\mathcal{L}_{seg}^{st}
\end{equation}
Here, we set $\lambda_{seg}^{st}=\lambda_{seg}^{pT}$ for a balanced segmentation objective between the annotated pseudo-target images and the pseudo-labeled original target images.

\section{Experiments and results}

\subsection{Experimental settings}

\subsubsection{Datasets}

\paragraph{\textbf{BraTS}}

We first evaluate MoDATTS on the BraTS 2020 challenge dataset \citep{b3,b2,b1}, adapted for the cross-modality brain tumor segmentation problem where images are known to be diseased (presence of tumors) or healthy (absence tumors). Amongst the 369 brain volumes available in BraTS, 37 volumes were allocated to each of the validation (10\%) and test (10\%) sets. The 295 volumes left were used for training (80\%). Based only on brain tissue, each volume was mean-centered, divided by five times the standard deviation and clipped to the [-1,1] interval. Datasets were then assembled from each distinct pair of the four MRI contrasts available (T1, T2, T1ce and FLAIR) for the modality adaptation task. To constitute unpaired training data, we used only one specific contrast (source or target) per training volume. We could therefore experiment with twelve different combinations of unpaired source/target modalities. Even though it is not clinically useful to learn cross-sequence segmentation if multi-parametric acquisitions are performed as is the case in BraTS, this modified version of the dataset provides an excellent study case to assess the actual performance of any modality adaptation method for tumor segmentation. Although the dataset offers several segmentation classes (enhancing tumor, peritumoral edema, necrotic and non-enhancing tumor core), note that we only consider the whole tumors as our segmentation objective.

\paragraph{\textbf{CrossMoDA}}

We also used the dataset from the 2022 CrossMoDA domain adaptation challenge \citep{crossmoda_1,crossmoda_2} for a cross-modality vestibular schwannoma segmentation task. The training dataset is composed of 210 contrast-enhanced T1-weighted MR volumes with pixel-level annotations and 210 unannotated unpaired high-resolution T2-weighted MR volumes. An additional test set of 64 unannotated hrT2 images was available for performance evaluation of the models on the data challenge platform. The images were equally acquired in 2 distinct centers, \textit{London} and \textit{Tilburg}, and showed different resolutions and sizes. To mitigate these disparities, all the 3D images were first resampled to a spacing of $0.41\times0.41\times1$. Then, to align the volumes and later facilitate the split into known healthy and diseased samples, we selected a hrT2 image as an atlas to perform inter and intra modality affine registrations. We used the Advanced Normalization Tools module \citep{ants}, and employed the mutual information loss for T1ce images and the normalized cross-correlation loss for hrT2 volumes. Images were then cropped to an ROI of $256\times256\times60$. Each volume was finally mean-centered, divided by five times the standard deviation and clipped to the [-1,1] interval.

\subsubsection{Tumor-aware cross-modality translation}

\paragraph{\textbf{2D slicing}} Due to resource constraints, prior to 3D segmentation, MoDATTS achieves 2D cross-modality translation to generate pseudo-target samples from the source modality. Therefore, CrossMoDA and BraTS volumes were respectively split into full $256\times256$ and $240\times240$ 2D slices before being fed to the modality translation network.

\paragraph{\textbf{Architecture}} For the architecture, we leverage the recent works on vision transformers \citep{vit}, which were shown to be suitable for translation tasks \citep{transformer_based} when combined with fully-convolutional discriminators. We exploit the TransUnet model \citep{transunet}, a powerful 2D U-shaped network for medical images that has shown great performance on several segmentation tasks. The architecture of our generators is based on the hybrid ``R50-ViT'' TransUnet configuration that combines a ResNet-50 and a ViT model with 12 transformer layers. The encoder backbones were pre-trained on ImageNet \citep{image_net}. For each of the two modality generators, the TransUnet decoder is duplicated producing a segmentation decoder with a sigmoid output activation and a translation decoder with a $tanh$ output activation. As for discriminators, we use a convolutional multi-scale architecture as in \cite{disc} that averages output values across several scales. Further details on the different layers are showed in Table \ref{disc}. We use leaky ReLU with a slope of 0.2 as the non-linear activation function.

\begin{table}[H]
\caption{Discriminator architecture used for the modality translation stage. $\mathbf{LN = }$ Layer Normalization, $\mathbf{LR = }$ Leaky ReLU activation, $\mathbf{C = }$ Convolution. The same architecture is used to train the diseased-healthy translation in the second stage, but kernels are expanded to 3D.}
\begin{center}
\vspace{-0.2cm}
\begin{tabular}{cccc}
\hline\hline
\multicolumn{4}{c}{\textbf{Discriminators}}\\
\hline\hline
Layer & Channels & Kernel size &  Stride  \\
\hline
C & 60 & $4\times4$ & 1\\
LN+LR+C & 60 & $4\times4$ & 2\\
LN+LR+C & 120 & $4\times4$ & 2\\
LN+LR+C & 240 & $4\times4$ & 2\\
LN+LR+C & 480 & $4\times4$ & 2\\
C & 1 & $1\times1$ & 1 \\
\hline
\end{tabular}
\label{disc}
\end{center}
\vspace{-0.5cm}
\end{table}

\begin{table*}[htp]
\caption{Architectures used for training at the segmentation stage. Except for the common decoder for which we add an extra convolution layer at the bottleneck with kernel size $1\times1\times1$ to map the common code back to the encoder output channel number, we use identical architectures for all decoders. Ch. sm. and Ch. sf. respectively refers to the number of channels for the semi-supervised and self-supervised variants. At each layer we show the number of convolution blocks (Conv.), and the number of B-MDH - Bidirectional Multi-Head Attention - blocks (Trans.) along with the number of heads for each of these blocks (Heads). $\mathbf{\dagger = 2\times}$ down-scaling with tri-linear interpolation and passing semantic maps to multi-scale fusion module. $\mathbf{\ddagger = 2\times}$ up-sampling with tri-linear interpolation and long skip connection from multi-scale fusion module. $\star$ indicates which layer is duplicated in the shared residual/segmentation decoder.}
\addtolength{\tabcolsep}{-3pt}
\begin{tabular}{ccccccc}
\hline\hline
\multicolumn{7}{c}{\textbf{Encoder}}\\
\hline\hline
 & Ch. sf. & Ch. sm. & Conv. & Trans. & Heads &  Kernel  \\
\hline
\multirow{2}{*}{\shortstack{Conv.\\Stem}}&16 & 32 & 1 & 0 & 0 & $3\times3\times3$  \\
& 32 & 64 & 2 & 0 & 0 & $\dagger\hspace{0.1cm}3\times3\times3$   \\\hdashline
\multirow{3}{*}{\shortstack{B-MDH\\blocks}} & 64 & 128 & 0 & 2 & 2 & $\dagger\hspace{0.1cm}3\times3\times3$  \\
 & 128 & 256 & 0 & 4 & 8 & $\dagger\hspace{0.1cm}3\times3\times3$   \\
& 256 & 320 & 0 & 6 & 10 & $\dagger\hspace{0.1cm}3\times3\times3$   \\
\hline
\end{tabular}
\hfill
\begin{tabular}{ccccccc}
\hline\hline
\multicolumn{7}{c}{\textbf{Decoders}}\\
\hline\hline
 & Ch. sf. & Ch. sm. & Conv. & Trans. & Heads &  Kernel  \\
\hline
\multirow{2}{*}{\shortstack{B-MDH\\blocks}}& 128 & 256 & 0 & 4 & 8 & $\ddagger\hspace{0.1cm}3\times3\times3$  \\
& 64 & 128 & 0 & 2 & 4 & $\ddagger\hspace{0.1cm} 3\times3\times3$   \\\hdashline
\multirow{3}{*}{\shortstack{Conv.\\Decoder}} & 32 & 64 & 2 & 0 & 0 & $\ddagger\hspace{0.1cm}3\times3\times3$  \\
& 16 & 32 & 2 & 0 & 0 & $\ddagger\hspace{0.1cm}3\times3\times3$   \\
& 1 & 1 & 1 & 0 & 0 & $\star\hspace{0.1cm}1\times1\times1$   \\
\hline
\end{tabular}
\label{archi_genseg}
\vspace{-0.2cm}
\end{table*}

\paragraph{\textbf{Training}} Our modality translation model was trained for 200 epochs. We used a batch size of 15 with the AMSGrad optimizer with $\beta_1=0.5$, $\beta_2=0.999$, and a learning rate of 0.0001. Pixel-level ground-truth annotations were provided only for the source modality slices. The same on-the-fly 2D data augmentation as in \cite{genseg} was applied. The following loss parameters, defined in section \ref{stage_1}, yielded great translation and preserved tumor appearance across modalities for both datasets : $\lambda_{adv}^{mod}=1$, $\lambda_{seg}^{mod}=1$ and $\lambda_{cyc}=10$.
\vspace{-0.2cm}
\paragraph{\textbf{Synthetic target data generation}}
Once the translation model was trained, 2D source slices were augmented into pseudo-target images using the last state of the translation model. The latter were then assembled back to constitute the synthetic target 3D volumes required for the segmentation stage.

\begin{figure}[H]
\centering
\includegraphics[width=0.48\textwidth]{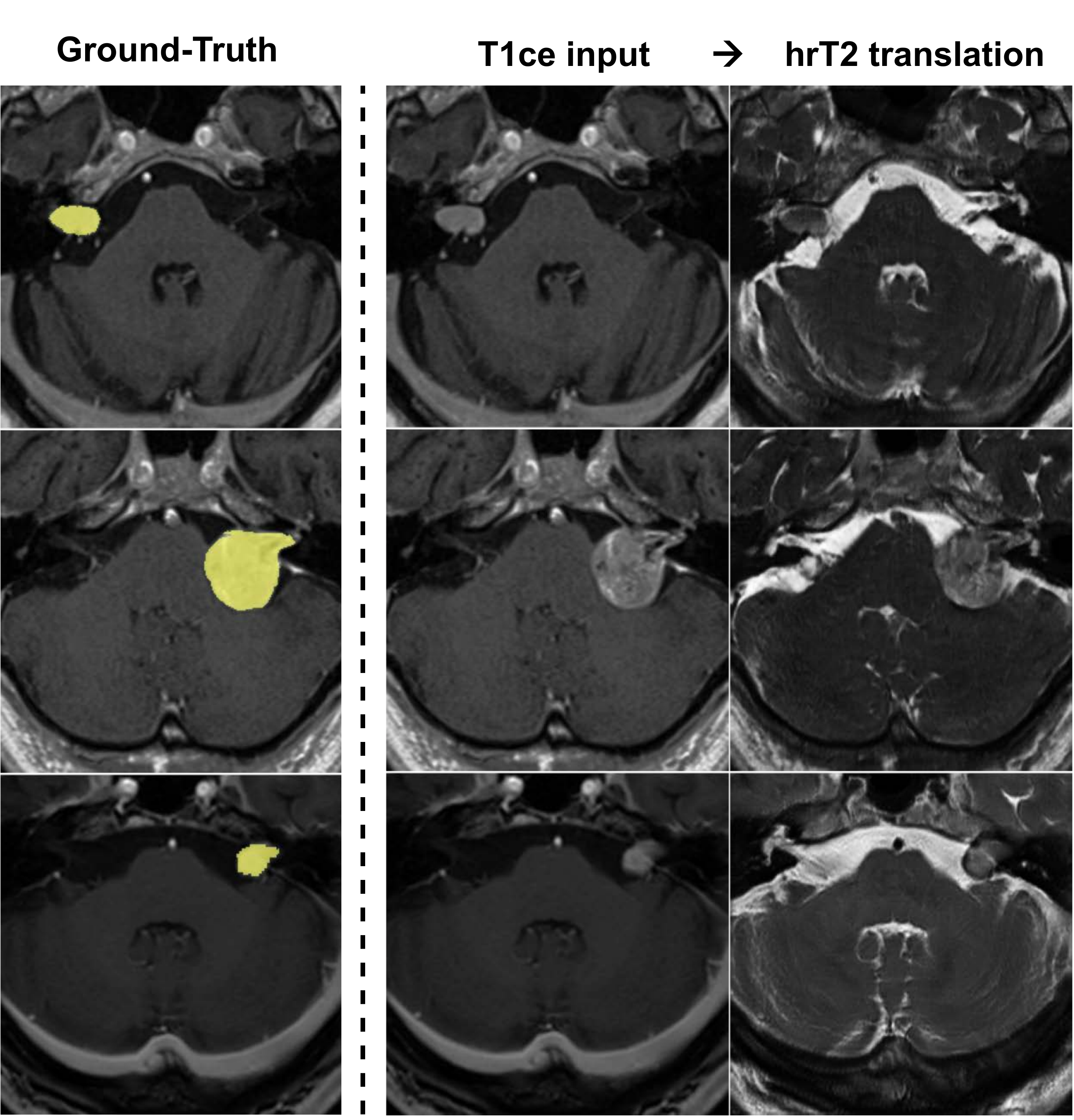}
\vspace{-0.5cm}
\caption{Cross-modality translation examples for the CrossMoDA dataset. We display several T1ce $\rightarrow$ hrT2 translations along with the VS segmentation ground truth. Tumor structures are preserved in the pseudo hrT2 images.}
\label{translation_crossmoda}
\end{figure}

\subsubsection{Semi-supervised target modality segmentation}

\paragraph{\textbf{Diseased/Healthy labeling}}
For the segmentation task, we created the sets of known healthy and diseased volumes. Each pseudo and real target 3D brain volumes were split into two hemispheres. For BraTS we attributed to each hemisphere the label P (presence of tumor) if any of its pixels was indicated to be part of a tumor by the ground truth segmentation, or the label A (absence of tumor) otherwise. Since Vestibular Schwannoma (VS) segmentations were available for T1ce in the CrossMoDA dataset, the same process was applied for the synthetized hrT2 images. However as the ground truth VS segmentations were not provided for original hrT2 data, manual labels were added to left or right hemispheres containing the VS for each of the 210 hrT2 volumes. We specify that in all the experiments the images are provided with absence/presence weak labels, distinct from the pixel-level annotations that we provided only to a subset of the data.

\begin{figure*}[htp]
\centering
\includegraphics[width=\textwidth]{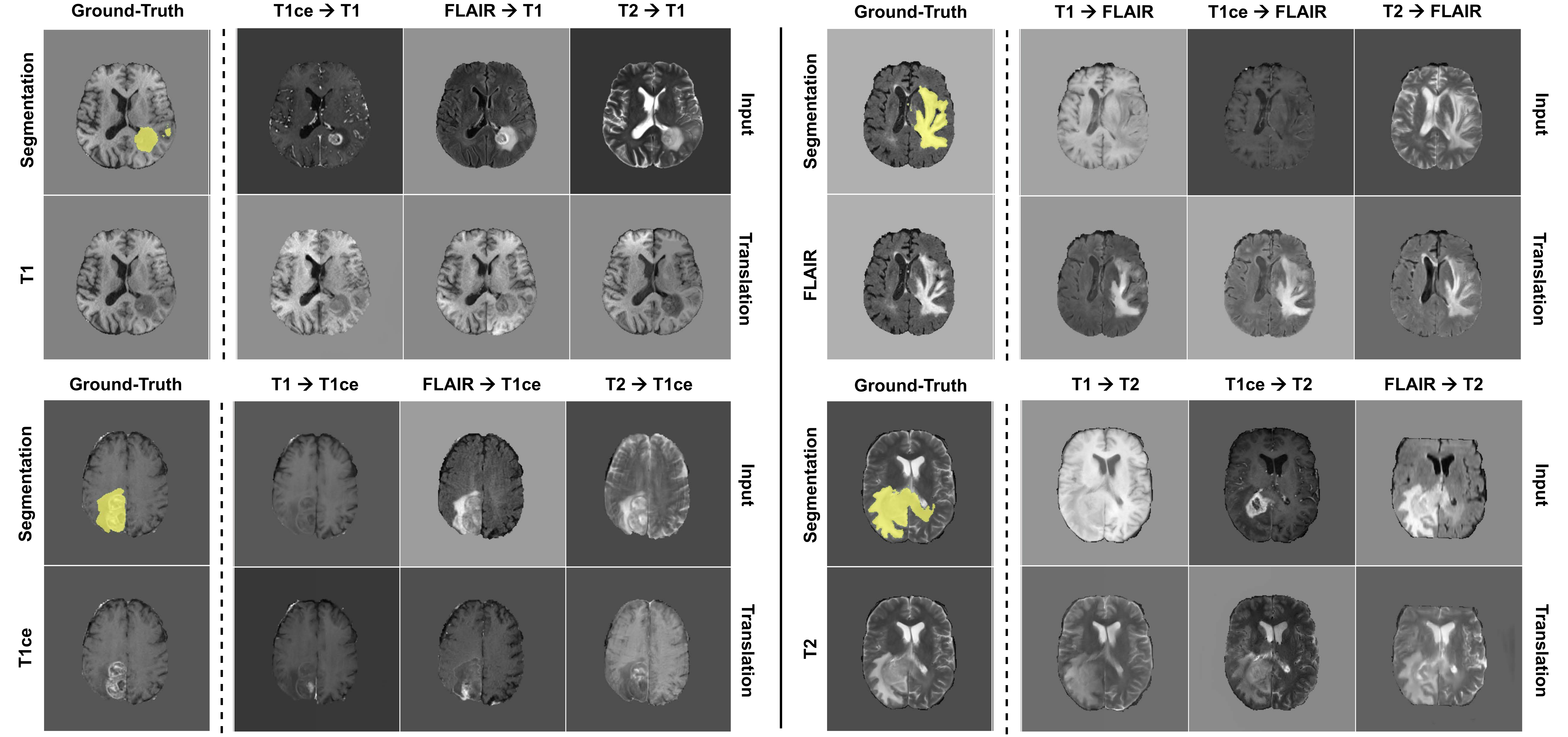}
\vspace{-0.5cm}
\caption{Cross-modality translation examples for the BraTS dataset. Each group of images represents all possible translations towards a specific modality (T1, T1ce, FLAIR or T2) for one case. For visual evaluation of the method we also show the corresponding ground truth target images and the whole tumor segmentations. The resulting visual appearances differ depending on the source modality, but tumor information is retained across all translations.}
\label{translation_brats}
\end{figure*}

\paragraph{\textbf{Architecture}}
Integrating the diseased/healthy translation to the domain adaptation framework requires a common and a residual decoder in addition to the standard segmentation encoder-decoder. As stated before, weights between the segmentation decoder and the residual decoder are shared so that segmentation is implicitly learned from the unsupervised objective. Therefore our model actually involves one unique residual/segmentation decoder but with two sets of normalization parameters. The latter also contains two distinct output layers, with tanh activation to generate residuals and sigmoid activation to yield segmentation maps. Our encoder and decoder architectures are based on a 3D Medformer \citep{medformer}, a recent data-scalable transformer architecture that outperformed the nnU-Net \citep{nnunet} and other vision transformers \citep{swinunetr,nnformer} on several medical image segmentation tasks. Note that we propose a self-supervised setup which only includes the supervision over pseudo-target samples and self-training (no common/residual decoders), and a semi-supervised setup which additionally performs the diseased/healthy translation on original target images. The number of weights per encoder/decoder for the semi-supervised variant had to be decreased in comparison to the self-supervised variant due to memory constraints. Further details on the different encoder and decoder layers are provided in Table \ref{archi_genseg}. In the semi-supervised variant, we also introduce two discriminators that are responsible for discriminating between real and generated diseased/healthy samples. Their architecture is the same as in the modality translation stage (cf Table \ref{disc}).

\begin{figure*}[!t]
\centering
\includegraphics[width=\textwidth]{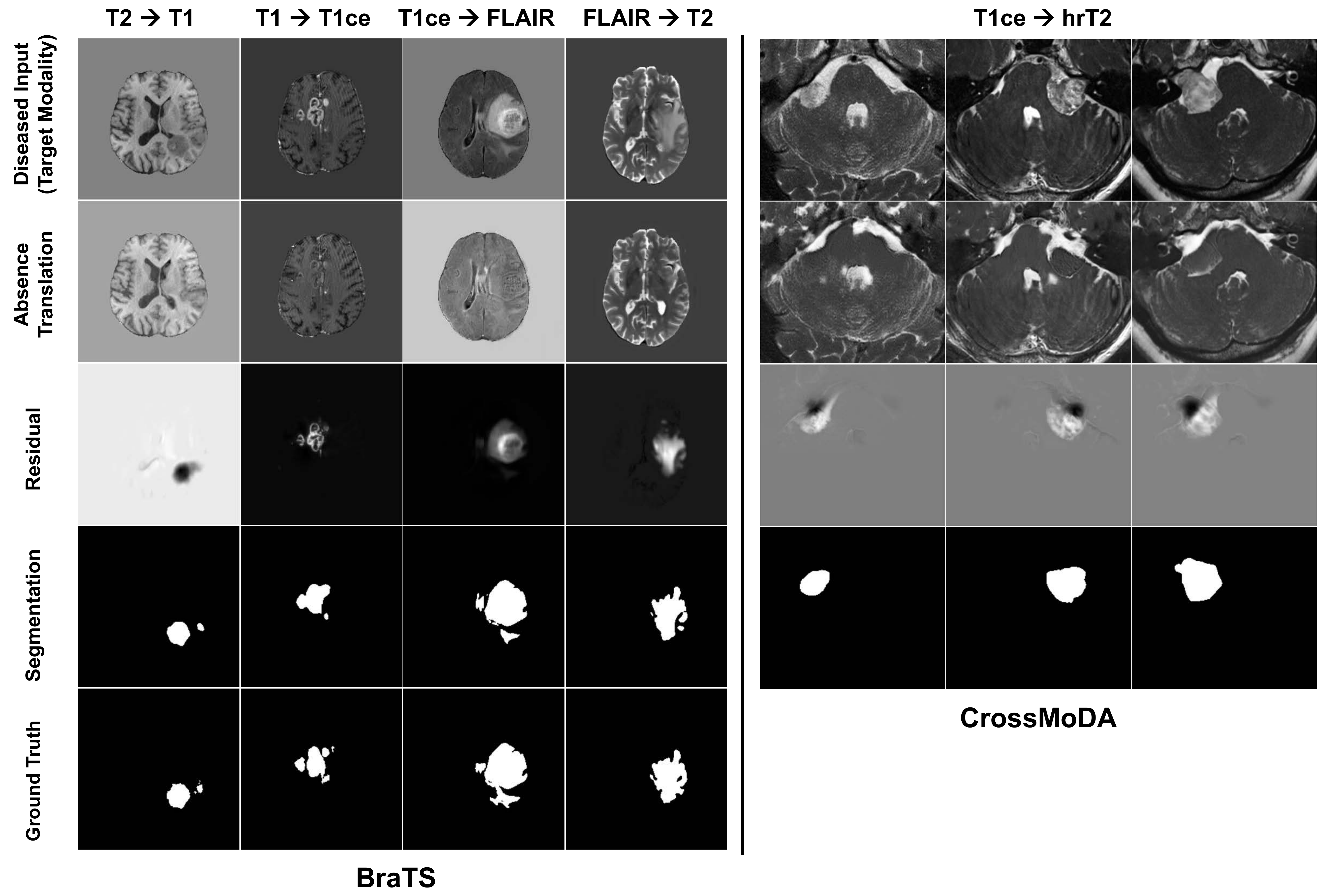}
\vspace{-0.5cm}
\caption{Examples of translations from Presence to Absence domains and resulting segmentation in a domain adaptation application where target modality had no pixel-level annotations provided. For BraTS, we show in each column a different source $\rightarrow$ target scenario. We do not display ground truth VS segmentations for CrossMoDA as hrT2 segmentation maps were not provided in the data challenge.}
\label{diseased-healthy}
\end{figure*}

\paragraph{\textbf{Training}} All segmentation models were trained for 300 epochs. We then performed three self-training iterations of 150 epochs each. For all runs, we applied 3D nnU-Net on-the-fly data augmentation and weights with the highest validation Dice score were saved for the next step. We used a batch-size of 2, and the same optimizer parameters as in the modality translation phase. The threshold $\alpha$ that defines the level of confidence required to keep the pseudo-labels was set to $0.6$ as in \cite{ne2e}. For BraTS, each training experiment was repeated three times, with a different random seed for weight initialization. We therefore report the mean of all test Dice scores with standard deviation across the three runs. Specifically in the segmentation stage, we performed 5-fold cross-validation on the training set for each CrossMoDA experiment. Ensembling was achieved with the resulting models for performance evaluation on the test dataset. As we were limited for quantitative evaluation on the online CrossMoDA data challenge platform, each experiment was evaluated only once. Note that for VS segmentation we applied up-sampling on large heterogeneous and small-sized tumors along with tumor intensity augmentation, as encouraged by \cite{LATIM}.

\vspace{-0.2cm}

\paragraph{\textbf{Hyper-parameter search}} Our approach involved the following strategy : (1) increasing the weights of the reconstruction terms $\lambda_{rec}$ and $\lambda_{lat}$ relatively to the adversarial term $\lambda_{adv}^{gen}$ until mode dropping stops occuring; and (2) subsequently determining the optimal weight $\lambda_{seg}^{pT}$ for supervision from the synthetic data. We found that the following parameters (normalized to equal 1) yielded great diseased/healthy translations for BraTS : $\lambda_{adv}^{gen}=5$, $\lambda_{rec}=50$, and $\lambda_{lat}=5$. For CrossMoDA, stronger reconstruction constraints were required as $\lambda_{rec}=75$ and $\lambda_{lat}=10$ yielded visually better results. In a standard domain adaptation scenario where 100\% of source data is provided with pixel-level annotations and all the target images are unnanotated, we found that $\lambda_{seg}^{pT}=100$ was optimal. Note that, during the first training of the segmentation model (prior to self-training), increasing $\lambda_{seg}^{pT}$ involves higher dependence on the synthetic target data. When lowering the number of samples provided with pixel-level annotations in the source modality, (1) the tumor appearances in the generated pseudo-target images are likely to be less accurate and (2) the segmentation model is fed with fewer annotated synthetic samples. This requires adjusting $\lambda_{seg}^{pT}$ down, so that the segmentation model relies more on the unsupervised tumor delineation objective rather than on the supervision from the synthetic data. When using $70\%$, $40\%$, $10\%$ and $1\%$ of the source modality annotations, we set $\lambda_{seg}^{pT}$ to respectively $50$, $25$, $1$ and $0.1$. These $\lambda_{seg}^{pT}$ values were tuned on CrossMoDA and reused for BraTS. 
\vspace{-0.1cm}

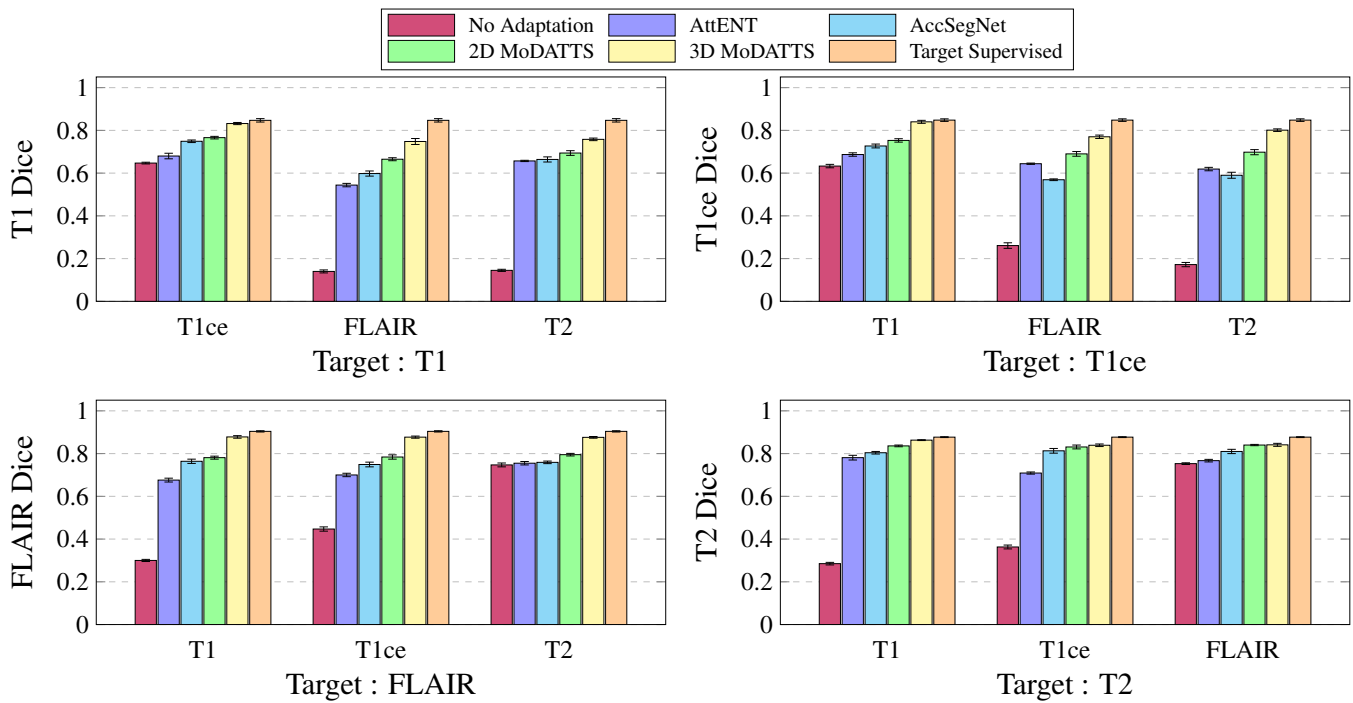
\begin{figure*}[!t]\centering
\subfloat{\resizebox{0.8\textwidth}{!}{\hspace{-1.9cm}\begin{tikzpicture}
    \begin{axis}[
        name={ax1},
        label style = {font=\Large},
        ticklabel style = {font=\large}, 
        width  = 0.6\textwidth,
        height = 5.3cm,
        ymax=1.05,
        major x tick style = transparent,
        ybar=2*\pgflinewidth,
        bar width=10pt,
        ymajorgrids = true,
        grid style=dashed,    
        ylabel={T1 Dice},
        xlabel={Target : T1},
        ytick={0,0.2,0.4,0.6,0.8,1},
        yticklabels={0, 0.2, 0.4, 0.6, 0.8, 1},
        symbolic x coords={T1ce, FLAIR, T2},
        xtick = data,
        scaled y ticks = false,
        enlarge x limits=0.3,
        xlabel style={at={(0.5,-1ex)}},
        ymin=0,
        legend cell align=left,
        legend image code/.code={%
                    \draw[#1, draw=black] (0.05cm,-0.08cm) rectangle (0.6cm,0.1cm);1
                },
        legend style={legend columns=3, anchor=south west,              nodes={scale=1, transform shape},
                at={(0.5,1.03)},
                column sep=1ex
        },
        legend image post style={scale=2}
    ]
        \addplot+[style={black,fill=purple!70,mark=none}, error bars/.cd, y dir=both, y explicit, error bar style=black]
            coordinates {(T1ce, 0.647) +- (0, 0.004)  (FLAIR, 0.140) +- (0, 0.007)  (T2,0.145) +- (0, 0.005) };
            
        \addplot[style={black,fill=blue!40,mark=none}, error bars/.cd, y dir=both, y explicit, error bar style=black]
            coordinates {(T1ce, 0.680) +- (0, 0.013)  (FLAIR, 0.544) +- (0, 0.008)  (T2,0.657) +- (0, 0.003) };
            
        \addplot[style={black,fill=cyan!40,mark=none}, error bars/.cd, y dir=both, y explicit, error bar style=black]
            coordinates {(T1ce, 0.749) +- (0,0.006)  (FLAIR, 0.598) +- (0, 0.012)  (T2,0.664) +- (0, 0.012) };
            
        \addplot[style={black,fill=green!40,mark=none}, error bars/.cd, y dir=both, y explicit, error bar style=black]
            coordinates {(T1ce, 0.766) +- (0, 0.006)  (FLAIR, 0.665) +- (0, 0.007)  (T2,0.694) +- (0, 0.011) };

        \addplot[style={black,fill=yellow!40,mark=none}, error bars/.cd, y dir=both, y explicit, error bar style=black]
            coordinates {(T1ce, 0.832) +- (0, 0.005)  (FLAIR, 0.748) +- (0, 0.014)  (T2,0.758) +- (0, 0.006) };        

        \addplot[style={black,fill=orange!40,mark=none}, error bars/.cd, y dir=both, y explicit, error bar style=black]
            coordinates {(T1ce, 0.847) +- (0, 0.008)  (FLAIR, 0.847) +- (0, 0.008)  (T2, 0.847) +- (0, 0.008) };
        \legend{No Adaptation, AttENT, AccSegNet, 2D MoDATTS, 3D MoDATTS, Target Supervised}   
            
    \end{axis}
    \hspace{1.9cm}
    \begin{axis}[
        at={(ax1.south east)},
        label style = {font=\Large},
        ticklabel style = {font=\large}, 
        width  = 0.6\textwidth,
        height = 5.3cm,
        ymax=1.05,
        major x tick style = transparent,
        ybar=2*\pgflinewidth,
        bar width=10pt,
        ymajorgrids = true,
        grid style=dashed,    
        ylabel={T1ce Dice},
        xlabel={Target : T1ce},
        ytick={0,0.2,0.4,0.6,0.8,1},
        yticklabels={0, 0.2, 0.4, 0.6, 0.8, 1},
        symbolic x coords={T1, FLAIR, T2},
        xtick = data,
        scaled y ticks = false,
        enlarge x limits=0.3,
        xlabel style={at={(0.5,-1ex)}},
        ymin=0,
        legend cell align=left,
        legend style={                nodes={scale=0.5, transform shape},
                at={(0.99,0.99)},
                column sep=1ex
        },
        legend image post style={scale=0.5}
    ]
        \addplot+[style={black,fill=purple!70,mark=none}, error bars/.cd, y dir=both, y explicit, error bar style=black]
            coordinates {(T1, 0.633) +- (0, 0.008)  (FLAIR, 0.261) +- (0, 0.013)  (T2,0.172) +- (0, 0.010) };
            
        \addplot[style={black,fill=blue!40,mark=none}, error bars/.cd, y dir=both, y explicit, error bar style=black]
            coordinates {(T1, 0.687) +- (0, 0.008)  (FLAIR, 0.644) +- (0, 0.003)  (T2,0.619) +- (0, 0.008) };
            
        \addplot[style={black,fill=cyan!40,mark=none}, error bars/.cd, y dir=both, y explicit, error bar style=black]
            coordinates {(T1, 0.727) +- (0,0.009)  (FLAIR, 0.569) +- (0, 0.004)  (T2,0.590) +- (0, 0.014) };
            
        \addplot[style={black,fill=green!40,mark=none}, error bars/.cd, y dir=both, y explicit, error bar style=black]
            coordinates {(T1, 0.753) +- (0, 0.008)  (FLAIR, 0.690) +- (0, 0.011)  (T2,0.698) +- (0, 0.012) };

        \addplot[style={black,fill=yellow!40,mark=none}, error bars/.cd, y dir=both, y explicit, error bar style=black]
            coordinates {(T1, 0.840) +- (0, 0.007)  (FLAIR, 0.770) +- (0, 0.008)  (T2,0.801) +- (0, 0.006) };        

        \addplot[style={black,fill=orange!40,mark=none}, error bars/.cd, y dir=both, y explicit, error bar style=black]
            coordinates {(T1, 0.848) +- (0, 0.006)  (FLAIR, 0.848) +- (0, 0.006)  (T2, 0.848) +- (0, 0.006) };
        
    \end{axis}
\end{tikzpicture}}}\\
\vspace{-0.3cm}
\subfloat{\resizebox{0.8\textwidth}{!}{\hspace{-1.9cm}\begin{tikzpicture}
    \begin{axis}[
        name={ax1},
        label style = {font=\Large},
        ticklabel style = {font=\large}, 
        width  = 0.6\textwidth,
        height = 5.3cm,
        ymax=1.05,
        major x tick style = transparent,
        ybar=2*\pgflinewidth,
        bar width=10pt,
        ymajorgrids = true,
        grid style=dashed,    
        ylabel={FLAIR Dice},
        xlabel={Target : FLAIR},
        ytick={0,0.2,0.4,0.6,0.8,1},
        yticklabels={0, 0.2, 0.4, 0.6, 0.8, 1},
        symbolic x coords={T1, T1ce, T2},
        xtick = data,
        scaled y ticks = false,
        enlarge x limits=0.3,
        xlabel style={at={(0.5,-1ex)}},        
        ymin=0,
        legend cell align=left,
        legend style={              nodes={scale=0.7, transform shape},
                at={(0.5,1.5)},
                column sep=1ex
        },
        legend image post style={scale=0.5}
    ]
        \addplot+[style={black,fill=purple!70,mark=none}, error bars/.cd, y dir=both, y explicit, error bar style=black]
            coordinates {(T1, 0.300) +- (0, 0.005)  (T1ce, 0.447) +- (0, 0.010)  (T2,0.747) +- (0, 0.009) };
            
        \addplot[style={black,fill=blue!40,mark=none}, error bars/.cd, y dir=both, y explicit, error bar style=black]
            coordinates {(T1, 0.676) +- (0, 0.009)  (T1ce, 0.700) +- (0, 0.008)  (T2,0.755) +- (0, 0.008) };
            
        \addplot[style={black,fill=cyan!40,mark=none}, error bars/.cd, y dir=both, y explicit, error bar style=black]
            coordinates {(T1, 0.764) +- (0,0.010)  (T1ce, 0.749) +- (0, 0.011)  (T2,0.759) +- (0, 0.006) };
            
        \addplot[style={black,fill=green!40,mark=none}, error bars/.cd, y dir=both, y explicit, error bar style=black]
            coordinates {(T1, 0.781) +- (0, 0.007)  (T1ce, 0.784) +- (0, 0.010)  (T2,0.795) +- (0, 0.006) };

        \addplot[style={black,fill=yellow!40,mark=none}, error bars/.cd, y dir=both, y explicit, error bar style=black]
            coordinates {(T1, 0.878) +- (0, 0.006)  (T1ce, 0.877) +- (0, 0.005)  (T2,0.876) +- (0, 0.004) };        

        \addplot[style={black,fill=orange!40,mark=none}, error bars/.cd, y dir=both, y explicit, error bar style=black]
            coordinates {(T1, 0.904) +- (0, 0.003)  (T1ce, 0.904) +- (0, 0.003)  (T2, 0.904) +- (0, 0.003) };
            
    \end{axis}
    \hspace{1.9cm}
    \begin{axis}[
        at={(ax1.south east)},
        label style = {font=\Large},
        ticklabel style = {font=\large}, 
        width  = 0.6\textwidth,
        height = 5.3cm,
        ymax=1.05,
        major x tick style = transparent,
        ybar=2*\pgflinewidth,
        bar width=10pt,
        ymajorgrids = true,
        grid style=dashed,    
        ylabel={T2 Dice},
        xlabel={Target : T2},
        ytick={0,0.2,0.4,0.6,0.8,1},
        yticklabels={0, 0.2, 0.4, 0.6, 0.8, 1},
        symbolic x coords={T1, T1ce, FLAIR},
        xtick = data,
        scaled y ticks = false,
        enlarge x limits=0.3,
        xlabel style={at={(0.5,-1ex)}},
        ymin=0,
        legend cell align=left,
        legend style={               nodes={scale=0.5, transform shape},
                at={(0.99,0.99)},
                column sep=1ex
        },
        legend image post style={scale=0.5}
    ]
        \addplot+[style={black,fill=purple!70,mark=none}, error bars/.cd, y dir=both, y explicit, error bar style=black]
            coordinates {(T1, 0.285) +- (0, 0.006)  (T1ce, 0.363) +- (0, 0.009)  (FLAIR,0.753) +- (0, 0.004) };
            
        \addplot[style={black,fill=blue!40,mark=none}, error bars/.cd, y dir=both, y explicit, error bar style=black]
            coordinates {(T1, 0.781) +- (0, 0.011)  (T1ce, 0.709) +- (0, 0.005)  (FLAIR,0.767) +- (0, 0.006) };
            
        \addplot[style={black,fill=cyan!40,mark=none}, error bars/.cd, y dir=both, y explicit, error bar style=black]
            coordinates {(T1, 0.804) +- (0,0.006)  (T1ce, 0.813) +- (0, 0.011)  (FLAIR,0.810) +- (0, 0.010) };
            
        \addplot[style={black,fill=green!40,mark=none}, error bars/.cd, y dir=both, y explicit, error bar style=black]
            coordinates {(T1, 0.836) +- (0, 0.004)  (T1ce, 0.831) +- (0, 0.009)  (FLAIR,0.840) +- (0, 0.003) };

        \addplot[style={black,fill=yellow!40,mark=none}, error bars/.cd, y dir=both, y explicit, error bar style=black]
            coordinates {(T1, 0.863) +- (0, 0.002)  (T1ce, 0.839) +- (0, 0.006)  (FLAIR,0.841) +- (0, 0.007) };        

        \addplot[style={black,fill=orange!40,mark=none}, error bars/.cd, y dir=both, y explicit, error bar style=black]
            coordinates {(T1, 0.877) +- (0, 0.002)  (T1ce, 0.877) +- (0, 0.002)  (FLAIR, 0.877) +- (0, 0.002) };
        
    \end{axis}
\end{tikzpicture}}}
\vspace{-0.3cm}
\caption{Dice performance on the target modality for each possible modality pair. Pixel-level annotations were only provided in the source modality indicated on the x axis. We compare results for MoDATTS (2D and 3D) with AccSegNet and AttENT domain adaptation baselines. We also show Dice scores for supervised segmentation models respectively trained with all annotations on source data (No adaptation) and on target data (Target supervised) as for lower and upper bounds of the cross-modality segmentation task.}
\label{raw dice}
\end{figure*}

\subsubsection{Model comparison}
When evaluating our model on BraTS, we compare the performance of the proposed approach against state-of-the-art domain-adaptive medical image segmentation models AccSegNet \citep{constrained} and AttEnt \citep{attent}. We used available GitHub code for the two baselines and performed fine-tuning on our data. Because these two methods are 2D,  for fair comparison, we also evaluate a 2D version of MoDATTS. Note that unless ``2D'' is specified, MoDATTS refers to the 3D version of our model. For CrossMoDA, we compare the performance of MoDATTS against the top 4 teams in the validation phase of the data challenge. The VS Dice scores and Average Symmetric Surface Distances (ASSD) for these methods were provided in the leader-board. For further experiments, the team Super-Poly \citep{Super-poly} was the only one to make its code available. Their MSF-nnU-Net model ranked $4^{th}$ in the data challenge but we believe their approach constitutes a reasonable baseline as it used a nnU-Net in the segmentation stage, similarly to the other top methods.
\vspace{-0.1cm}

\subsection{Domain adaptation}

The first set of experiments consists in evaluating MoDATTS in a standard domain adaptation scenario where all of the source data is provided with pixel-level annotations and all the target images are unannotated. This is the standard scenario for CrossMoDA as hrT2 segmentations are not available. As for BraTS we drew all possible source and target modality pairs from T1, T2, FLAIR and T1ce, and pixel-level annotations were only retained for the source modality.

\vspace{-0.2cm}
\paragraph{\textbf{Qualitative results}} We show in Fig. \ref{translation_brats} several generated samples of pseudo-target brain images after training of the modality translation model, when all the source samples were provided with pixel-level annotations. Interestingly, each source modality leaves its own style footprint in the generated images as the tumor appearances in the resulting target translations differ accordingly. An interesting feature of our model is that the tumor structures seem to be visually preserved across the modality translation. For instance, note that the different substructures of the tumor can still be differentiated in the FLAIR $\rightarrow$ T1 modality translation. Also notice that the translation model can successfully augment hypo-intense tumors that are hardly distinguishable from the background (e.g. T1ce $\rightarrow$ FLAIR or T1 $\rightarrow$ FLAIR). Similarly we show in Fig. \ref{translation_crossmoda} several T1ce $\rightarrow$ hrT2 VS translations. This further proves successful maintenance of tumor layouts during the pseudo-target sample generations, even for small lesions.

An illustration of several translations from the diseased to the healthy domain for the brain tumor and VS segmentation tasks are displayed in Fig. \ref{diseased-healthy}. As depicted in the figure, even without any pixel-level annotations for the target modality, the tumors were effectively separated from the brain, leading to a successful translation from the presence to the absence domain, as well as accurate segmentation. It is worth noting that even for lesions appearing hypo-intense, as in brain T1 and T1ce sequences, MoDATTS can effectively handle complex residuals and alternatively convert them into reliable segmentation results.
 \vspace{-0.2cm}
\paragraph{\textbf{Quantitative results}} We present the resulting absolute Dice scores for MoDATTS and each evaluated baseline on the brain tumor dataset in Fig. \ref{raw dice}. Note that the results for MoDATTS correspond to the self-supervised variant as it was more effective than the semi-supervised variant when 100\% of the source data was annotated (see section \ref{annotation_deficit}). Additionally we display the results obtained by a supervised Medformer model with the same backbone architecture as the self-supervised variant of MoDATTS, trained on the one hand with source data without any domain adaptation strategy and on the other hand with fully annotated target data, which respectively act as lower and upper bounds of the domain adaptation task. As expected, models without domain adaptation approaches trained on the source data fail to properly segment tumors on the target modality, particularly for modality pairs that show high domain shifts (e.g. T1/FLAIR or T1ce/T2). Note that MoDATTS shows great performance as it outperforms AttENT and AccSegNet with a considerable margin on the target modality. On average, over the 12 different domain adaptation experiments we report that 3D MoDATTS reaches $95\%$ of the target supervised model performance. These results demonstrate that our transformer-based modality translation approach is effective and is able to produce reliable pseudo-target images to train a segmentation model to delineate tumors in the target modality. In comparison AttENT and AccSegNet  reached $79\%$ and $82\%$, respectively. The 2D version of MoDATTS, reached $87\%$ of the target supervised model performance. This demonstrates that the superior performance of our model is not solely attributable to working in 3D.

We also report in Table \ref{Crossmoda_res} the VS Dice scores and ASSD on the CrossMoDA dataset. MoDATTS shows similar VS segmentation performance as team LaTIM who ranked first in the data challenge, and outperforms runner-up entries. This further proves that our method is effectively able to reduce the performance gap due to domain shifts in cross-modality tumor segmentation. Although the performance gains are limited, note that these approaches were specifically designed to perform on the CrossMoDA challenge and may not generalize well to other datasets. In contrast, our results on BraTS and CrossMoDA indicate that MoDATTS is competitive in several domain adaptation tumor segmentation tasks. We further note that the only competing approach (LaTIM) requires training a SinGAN \citep{singan} purposely adapted for CrossMoDA to augment and diversify the target VS appearances, in addition to the conventional modality translation and segmentation models. MoDATTS also achieves such data augmentation through the healthy $\rightarrow$ diseased translation objective. However it is encompassed in the segmentation model, therefore mitigating the need for an additional step.

\begin{table}[h]
\renewcommand\thetable{3}
\centering
\parbox{\linewidth}{
\caption{Dice score and ASSD for the VS segmentation on the target hrT2 modality in the CrossMoDA challenge. We compare our performance with the top 4 teams in the validation phase. The standard deviations reported correspond to the performance variation across the 64 test cases.}
\vspace{-0.2cm}
\label{Crossmoda_res}
\resizebox{0.48\textwidth}{!}{
\begin{tabular}{ccc}
\firsthline
Model &  VS Dice $\uparrow$ & VS ASSD $\downarrow$ \\
\hline
\citep{ne2e} ne2e  & $0.847\pm0.063$ & $0.551\pm0.303$ \\
\citep{Super-poly} Super-Poly  & $0.849\pm0.068$   & $0.520\pm0.229$ \\
\citep{MAI} MAI  & $0.852\pm0.089$ &   $0.475\pm0.207$\\
\citep{LATIM} LaTIM & $0.868\pm0.060$  & $\mathbf{0.430\pm0.178}$\\
MoDATTS (Ours)  & $\mathbf{0.870\pm0.048}$  & $0.432\pm0.175$ \\
\lasthline
\end{tabular}}}
\vspace{-0.6cm}
\end{table}

\begin{table*}[htp]\centering
\parbox{0.73\linewidth}{
\caption{Brain tumor segmentation Dice scores when using reference annotations for 100\% of source T2 data and various fractions (0\%, 10\%, 30\% and 50\%) of target T1ce data during training. We also show the \% of the target supervised model performance (TSMP) that is reached by MoDATTS for the corresponding fractions of T1ce annotations.}
\vspace{-0.2cm}
\centering
    \resizebox{\linewidth}{!}{
\begin{tabular}{c|c|c|c|c|c}
T1ce annotations & 0\% & 10\% & 20\% & 30\% & 50\% \\
\hline
T1ce Dice Score & $0.801\pm0.006$ & $0.826\pm0.006$ & $0.839\pm0.005$
 & $0.844\pm0.007$
 & $0.847\pm0.005
$\\\hdashline
\% of TSMP & 94.5\% & 97.5\% & 99\% & 99.5\% & 100\%\\
\end{tabular}}
\label{reach}}
\end{table*}
\vspace{0.4cm}
\begin{figure*}[!t]\centering
\captionsetup[subfigure]{oneside,margin={2.2cm,0cm}}
\subfloat[BraTS Dataset]{\resizebox{0.575\textwidth}{!}{\begin{tikzpicture}
\begin{axis}[
    name=ax1,
    title style = {font=\huge},
    label style = {font=\Large},
    ticklabel style = {font=\Large},
    xlabel={Source T1 annotations},
    ylabel={Target T2 Dice Score},
    ymin=0.38, ymax=0.92,
    ytick={0.4,  0.5,  0.6,  0.7,  0.8, 0.9},
    xtick={0.01, 0.2, 0.4, 0.6, 0.8, 1},
    xticklabels={1\%, 20\%, 40\%, 60\%, 80\%, 100\%},
        legend style={legend columns=2, cells={align=center},at={(1.13,1.06)},
     anchor=south},
    ymajorgrids=true,
    grid style=dashed
]
        \addplot[style={orange,mark=star, mark options={scale=1.5,fill=white}}]
             coordinates {(0.01, 0.770) (0.1,0.815) (0.4,0.844) (0.7,0.852) (1,0.857)};

        \addplot[style={ggreen,mark=triangle*, mark options={scale=2,fill=white}}]
             coordinates {(0.01, 0.423) (0.1,0.736) (0.4,0.782) (0.7,0.796) (1,0.804)};

        \addplot[style={ppurple,mark=star, mark options={scale=1.5,fill=white}}]
             coordinates {(0.01, 0.658) (0.1,0.794) (0.4,0.832) (0.7,0.848) (1,0.863)};
             
        \addplot[style={bittersweet,mark=*, mark options={scale=1.5,fill=white}}]
             coordinates {(0.01, 0.478) (0.1,0.674) (0.4,0.721) (0.7,0.747) (1,0.781)};
             
        \addplot[style={bblue,mark=diamond*, mark options={scale=2,fill=white}}]
            coordinates {(0.01, 0.734) (0.1, 0.778) (0.4,0.823) (0.7,0.829) (1,0.836)};

        \legend{Semi-supervised MoDATTS, AccSegNet, Self-supervised MoDATTS, AttENT, Semi-supervised 2D MoDATTS}
             
\end{axis}
\hspace{1.9cm}
\begin{axis}[
    at={(ax1.south east)},
    title style = {font=\huge},
    label style = {font=\Large},
    ticklabel style = {font=\Large},
    xlabel={Source T2 annotations},
    ylabel={Target T1 Dice Score},
    ymin=0.38, ymax=0.82,    
    xtick={0.01, 0.2, 0.4, 0.6, 0.8, 1},
    xticklabels={1\%, 20\%, 40\%, 60\%, 80\%, 100\%},
    ytick={0.1,0.2,0.3,0.4,0.5,0.6,0.7,0.8},    
    legend style={at={(1.03,0.5)},anchor=west},
    ymajorgrids=true,
    grid style=dashed
]
        \addplot[style={orange,mark=star, mark options={scale=1.5,fill=white}}]
             coordinates {(0.01, 0.684) (0.1,0.724) (0.4,0.743) (0.7,0.745) (1,0.748)};

        \addplot[style={ggreen,mark=triangle*, mark options={scale=2,fill=white}}]
             coordinates {(0.01, 0.43) (0.1,0.606) (0.4,0.643) (0.7,0.654) (1,0.664)};

        \addplot[style={ppurple,mark=star, mark options={scale=1.5,fill=white}}]
             coordinates {(0.01, 0.618) (0.1,0.703) (0.4,0.737) (0.7,0.753) (1,0.758)};

        \addplot[style={bittersweet,mark=*, mark options={scale=1.5,fill=white}}]
             coordinates {(0.01, 0.522) (0.1,0.635) (0.4,0.645) (0.7,0.65) (1,0.656)};
             
        \addplot[style={bblue,mark=diamond*, mark options={scale=2,fill=white}}]
            coordinates {(0.01, 0.582) (0.1, 0.671) (0.4,0.677) (0.7,0.685) (1,0.694)};

\end{axis}
\end{tikzpicture}}}
\hfill
\captionsetup[subfigure]{oneside,margin={1cm,0cm}}
\subfloat[CrossMoDA Dataset]{\resizebox{0.32\textwidth}{!}{\begin{tikzpicture}
\begin{axis}[
    title style = {font=\huge},
    label style = {font=\Large},
    ticklabel style = {font=\Large},
    xlabel={Source T1ce annotations},
    ylabel={Target hrT2 Dice Score},
        ymin=0.38, ymax=0.92,
    ytick={0.4, 0.5, 0.6,0.7,0.8,0.9},
    xtick={0.01, 0.2, 0.4, 0.6, 0.8, 1},
    xticklabels={1\%, 20\%, 40\%, 60\%, 80\%, 100\%},
    legend style={legend columns=1, cells={align=center},at={(0.5,1.02)},
     anchor=south},
    ymajorgrids=true,
    grid style=dashed
]
        \addplot[style={orange,mark=star, mark options={scale=1.5,fill=white}}]
            coordinates {(0.01, 0.7275) (0.1, 0.7932) (0.4, 0.8263) (0.7, 0.8378) (1,0.8512)};
            
        \addplot[style={ppurple,mark=star, mark options={scale=1.5,fill=white}}]
             coordinates {(0.01, 0.6217) (0.1, 0.7467) (0.4, 0.8194) (0.7, 0.8461) (1, 0.8700)};
             
        \addplot[style={bittersweet,mark=*, mark options={scale=1.5,fill=white}}]
             coordinates {(0.01, 0.4737) (0.1, 0.6934) (0.4, 0.7874) (0.7, 0.8206) (1,0.8436)};

         \legend{Semi-supervised MoDATTS, Self-supervised MoDATTS, MSF-nnU-Net}             
\end{axis}
\end{tikzpicture}}}
\caption{Dice scores when using reference annotations for 0\% of target data and various fractions (1\%, 10\%, 40\%, 70\% and 100\%) of source data during training. For BraTS, we picked the T1/T2 modality pair and ran the experiments in both T1 $\rightarrow$ T2 and T2 $\rightarrow$ T1 directions. For CrossMoDA, hrT2 annotations are not available so the experiment is run only in the T1ce $\rightarrow$ hrT2 direction. While performance is dropping at low \% of annotations for the baselines, semi-supervised MoDATTS shows in comparison only a slight decrease. For readability, standard deviations across the runs are not shown.}
\label{semi}
\end{figure*}
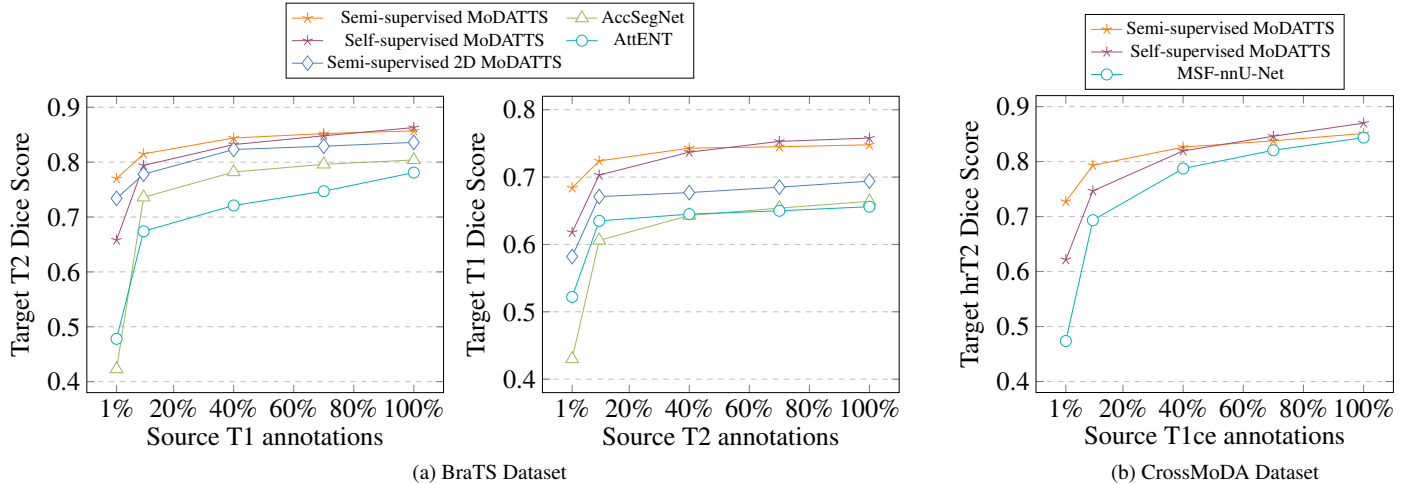

\subsection{Reaching supervised performance}

As mentionned in the previous section, MoDATTS performs well when the target modality is completely unannotated, as on average $95\%$ of the target supervised model performance is reached on BraTS. With the aim of determining the fraction of target modality annotations required to match the performance of a target supervised model, we trained MoDATTS (self-supervised) with a fully annotated source modality and increasing fractions of target annotations (0\%, 10\%, 20\%, 30\% and 50\%) on the BraTS T2 $\rightarrow$ T1ce domain adaptation task. Results are provided in Table \ref{reach}. We show that with a fully annotated source modality, it is sufficient to annotate 20\% of the target modality to reach 99\% (T1ce : $0.839\pm0.005$) of the target supervised model performance (T1ce : $0.848\pm0.006$). This emphasizes that the annotation burden could be reduced with our approach.

\subsection{Semi-supervision and annotation deficit}
\label{annotation_deficit}
\vspace{0.1cm}
MoDATTS introduces the ability to train with limited pixel-level annotations available in the source modality, a distinct feature over previous baselines. We show in Fig. \ref{semi} the Dice scores for models trained when $1\%$, $10\%$, $40\%$, $70\%$ or $100\%$ of the source modality's annotations were available combined with $0\%$ for the target modality. Note that the modality translation networks were retrained accordingly. While the performance of the baselines and the self-supervised variant of MoDATTS show a significant drop with fewer source annotations, the semi-supervised variant exhibits consistent performance with only slight degradation. Notably, the semi-supervised variant outperforms the self-supervised variant when less than $40\%$ of the source samples are annotated. For instance for BraTS T1$\rightarrow$T2 domain adaptation, semi-supervised MoDATTS with 1\% source annotations still reaches 88\% of the performance of a target (T2) supervised model, while the self-supervised variant only achieves 75\%. These findings validate that MoDATTS has the potential achieve robust performance even with a small fraction of annotated source images.

Note that when most of the source data is annotated, the performance gap between the self-supervised and semi-supervised variants remains small. When $100\%$ of source data is annotated, we report (self-supervised vs semi-supervised) target modality Dice scores of $0.863$ vs $0.857$ for T1 $\rightarrow$ T2 (BraTS), $0.758$ vs $0.748$ for T2 $\rightarrow$ T1 (BraTS), and $0.870$ vs $0.851$ for T1ce $\rightarrow$ hrT2 (CrossMoDA). This indicates that supervision from highly reliable synthetic data combined with self-training provide enough information to the segmentation model to close the domain gap. Finally, this small gap may be closed or reduced with access to better hardware since we had to reduce the size of the segmentation encoder-decoder in the semi-supervised variant from 38 million parameters (self-supervised) to 8 million parameters (semi-supervised).

\begin{table*}[!t]
\caption{Ablation studies : absolute Dice scores on the target modality. For BraTS, we selected T1 and T2 to be respectively the source and target modalities for these experiments. Ablations of the tumor-aware modality translation (TAMT) and self-training (ST) were performed when 100$\%$ of the source data was annotated with the self-supervised variant of MoDATTS, as it yielded the best performance. Alternatively, the ablations related to the diseased-healthy translation were performed on the semi-supervised variant when 1$\%$ of the source data was annotated. We also report for BraTS the \% of a target supervised model performance that is reached by the model (values indicated in parenthesis). Note that for CrossMoDA the standard deviations reported correspond to the performance variation across the 64 test cases.}
\centering
\begin{tabular}{clcc}
\firsthline
Source annotations & Ablations & BraTS : T1 $\rightarrow$ T2 & CrossMoDA : T1ce $\rightarrow$ hrT2 \\
\hline
\multirow{3}{*}{\shortstack{100\%\\(Self-supervised Variant)}} & \tikzcircle[fill=purple!70]{3pt} w/o TAMT w/o ST & $0.817\pm0.005$ (93\%) & $0.826\pm0.100$ \\
& \tikzcircle[fill=blue!40]{3pt} w/o ST & $0.844\pm0.001$ (96\%) & $0.851\pm0.051$ \\
& \tikzcircle[fill=cyan!40]{3pt} \textbf{Proposed (Self-sup.)} & $\mathbf{0.863\pm0.002}$ (98\%) & $\mathbf{0.870\pm0.048}$  \\\hdashline
\multirow{4}{*}{\shortstack{1\%\\(Semi-supervised Variant)}} & \tikzcircle[fill=green!40]{3pt} w/o P$\rightarrow$A and A$\rightarrow$P & $0.658\pm0.009$ (75\%) & $0.621\pm0.269$ \\
& \tikzcircle[fill=yellow!40]{3pt} w/o A $\rightarrow$ P & $0.715\pm0.012$ (82\%) & $0.666\pm0.222$ \\
& \tikzcircle[fill=orange!40]{3pt} w/o dual-use Res./Seg. & $0.739\pm0.008$ (84\%) & $0.707\pm0.177$ \\
& \tikzcircle[fill=red!40]{3pt} \textbf{Proposed (Semi-sup.)} &  $\mathbf{0.770\pm0.005}$ (88\%) & $\mathbf{0.727\pm 0.173}$ \\
\lasthline
\end{tabular}
\label{ablation}
\vspace{0.3cm}
\end{table*}
\begin{figure*}[!t]
\centering
\includegraphics[width=\textwidth]{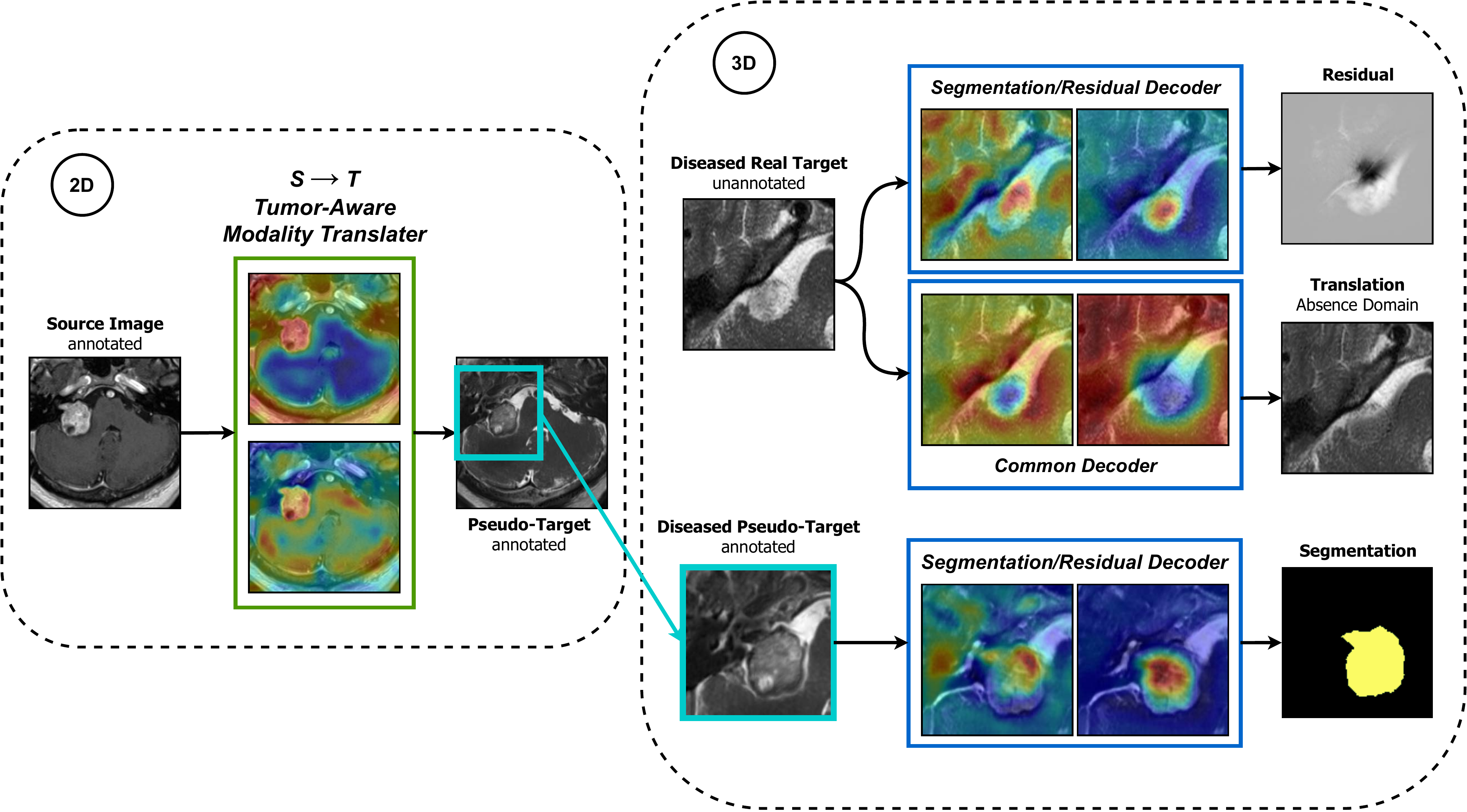}
\vspace{-0.5cm}
\caption{Attention maps yielded by the most confident transformer heads in MoDATTS. Red color indicates areas of focus while dark blue corresponds to locations ignored by the network. Note that the maps presented in the modality translation stage are produced by the encoder of the network as the decoder is fully convolutional and does not contain transformer blocks.}
\label{attention}
\end{figure*}

\subsection{Attention in MoDATTS}

Attention-based networks like transformers allow us to interrogate the model by analyzing the learned attention mechanisms. As suggested by \cite{confidence}, we computed the ``confidence'' of each attention head in the model as the average of its maximum attention weight. We show in Fig. \ref{attention} the attention maps generated by the most confident heads in MoDATTS on CrossMoDA. A confident head can be interpreted as one that assigns high attention values to specific regions of the image. We note that the transformer component (encoder) in the modality translation network tends to focus on global anatomical details of the image. Interestingly, it is also highlighting the VS, which emphasizes its ability to preserve tumor structures during the pseudo-target image synthesis. In the segmentation phase, the heads of the common and residual/segmentation decoders have different behaviors. Interestingly, the common decoder seems to avoid the tumor location, as a way to focus on the anatomical and healthy content of the image. As expected, the joint residual/segmentation decoder focuses on tumor areas in order to generate accurate residuals and segmentation maps. It also looks beyond the tumor. This is not surprising because the network has to compare the tumor to background tissue; also, a tumor can impact surrounding structures and the way the tumor appears in the image depends on the rest of the tissue.

\subsection{Ablation studies}

In the scenario where all the source samples have pixel-level annotations and none are available in the target modality, we conduct the following ablation experiments and report the results in Table \ref{ablation}. Specifically, we focus here on the self-supervised variant of MoDATTS (\tikzcircle[fill=cyan!40]{3pt}), as it exhibited the highest segmentation performance in this particular setup. Note that we chose T1 and T2 to be respectively the source and target modalities for the ablations on BraTS.

\vspace{-0.2cm}

\begin{figure}[!t]
\centering
\includegraphics[width=0.48\textwidth]{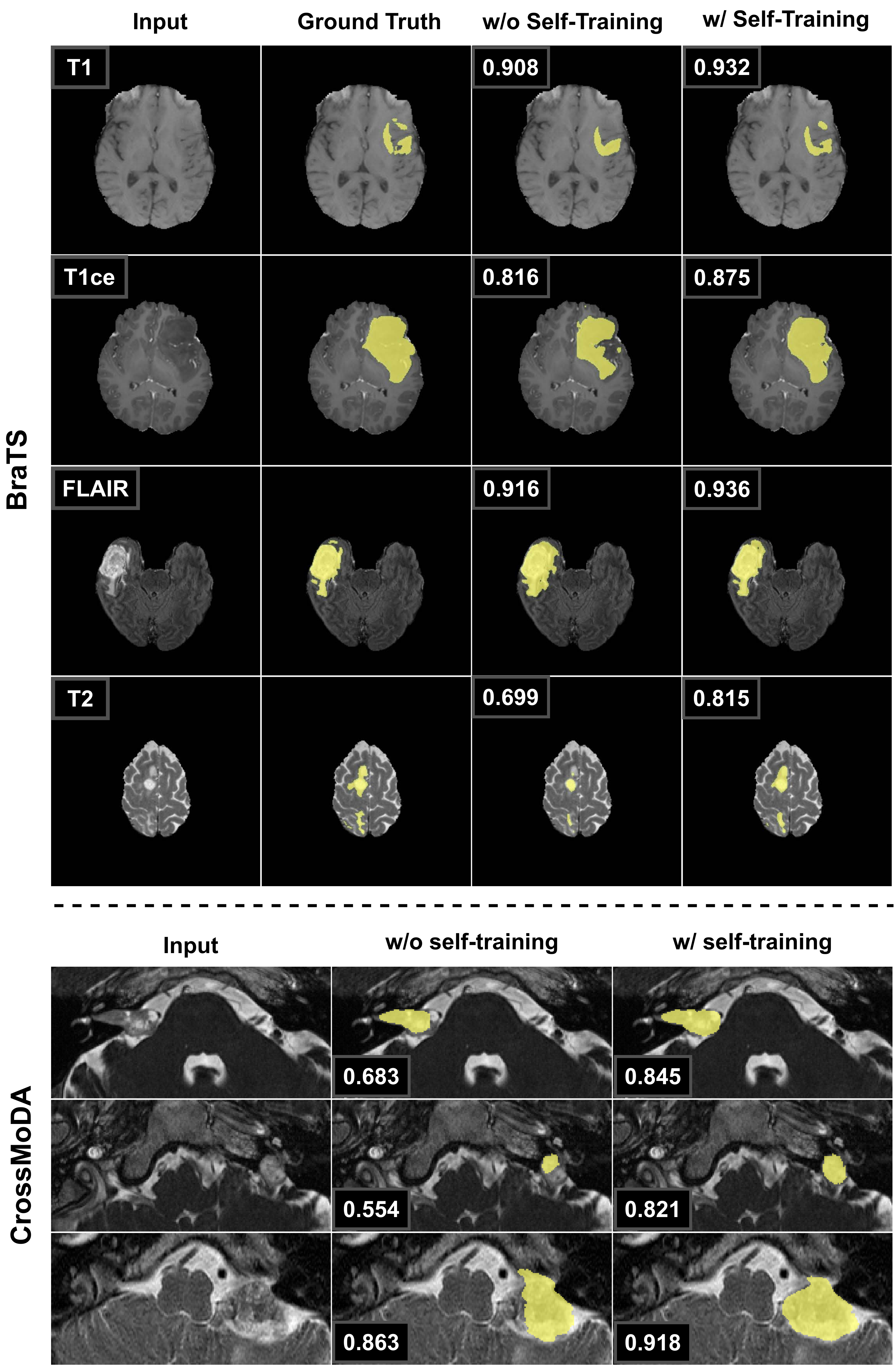}
\caption{Qualitative evaluation of the impact of self-training on the test set for brain tumor and VS segmentation tasks. Last two columns show, respectively, the segmentation map for the same model before and after the self-training iterations. For BraTS, each row illustrates a different scenario where the target modality - indicated in the first column - was not provided with any annotations. We also show ground truth tumor segmentations to visually assess the improvements of the model when self-training is performed, along with the dice score obtained on the whole volume. For CrossMoDA the ground truth VS segmentations on target hrT2 MRIs were not provided. Self-training helps to better leverage unnanotated data in the target modality and acts as a refining of the segmentation maps, which helps to reach better performance.}
\label{st}
\vspace{-0.3cm}
\end{figure}

\paragraph{\textbf{Self-training}} We evaluate the performance of MoDATTS before performing iterative self-training (\tikzcircle[fill=blue!40]{3pt}). We notice an improvement of around $+0.02$ Dice score in the segmentation performance after the process for VS and Brain tumor segmentation. Qualitative evaluation of the impact of self-training in the domain adaptation task for BraTS and CrossMoDA datasets is also provided in Fig. \ref{st}. The latter shows that the tumors on the test set are either filled or refined after self-training.

\vspace{-0.2cm}

\paragraph{\textbf{Tumor-aware modality translation}} To assess the value of additional tumor supervision in the modality translation stage, we retrained the translation model with $\lambda_{seg}^{mod} = 0$. Iterative self-training was not applied in the segmentation stage (\tikzcircle[fill=purple!70]{3pt}). As expected, the target modality segmentation performance dropped as compared to the previous ablation (BraTS : $-0.027$ and CrossMoDA : $-0.025$ in Dice). This implies that the joint tumor supervision in the modality translation stage actually helps to retain detailed lesion structures and provide more accurate pseudo-target images.\vspace{-0.3cm}
\\\\
The next ablations focus on the semi-supervised variant of MoDATTS (\tikzcircle[fill=red!40]{3pt}) and the contribution of the diseased-healthy translation when few annotated source samples are provided. We experiment with $1\%$ of annotated source data, as it is where the semi-supervised variant is the most relevant. Values are reported in Table \ref{ablation}, and interpretations are provided below.

\vspace{-0.2cm}
\paragraph{\textbf{Image-level supervision}}
We notice that only training the translation from diseased to healthy domains (P $\rightarrow$ A) suffices (\tikzcircle[fill=yellow!40]{3pt}), as 82\% of a target supervised model performance is reached on BraTS. But teaching the model to perform healthy to diseased (A $\rightarrow$ P) yields better performance (BraTS : $+0.055$ and CrossMoDA : +$0.061$ in Dice) by making more efficient use of the data. Note that when the whole diseased-healthy unsupervised objective is removed (\tikzcircle[fill=green!40]{3pt}), the segmentation performance on the target modality is on par with the one achieved by the self-supervised variant.

\vspace{-0.2cm}
\paragraph{\textbf{Separate residual and segmentation decoders}} Finally, we explored the effect of the decoders by employing a separate segmentation decoder instead of sharing the residual and segmentation weights (\tikzcircle[fill=orange!40]{3pt}). This separate version shows lower performance on the brain ($-0.31$ Dice) and VS ($-0.20$ Dice) datasets. This observation emphasizes that disentangling tumors from the background to perform diseased to healthy translations is similar to a segmentation objective, and therefore is beneficial for accurate cross-modality tumor segmentation when few source samples are annotated.

\section{Applications and extensions}

We have introduced a domain adaptation method to segment tumors on unnanotated target modality datasets from annotated or partially annotated source modality images. We have demonstrated the competitiveness and robustness of MoDATTS on cross-modality brain tumor and vestibular schwannoma MR sequences.

\paragraph{\textbf{Self-supervised vs semi-supervised variant}} The proposed model offers (1) a self-supervised variant that achieves supervision over pseudo-target samples and self-training; and (2) a semi-supervised variant that further includes real target modality images provided with diseased or healthy labels through unsupervised tumor disentanglement. Training the semi-supervised model requires more memory and expensive computation.  Due to our limited resources, the semi-supervised model is equipped with fewer parameters. As a consequence, when enough pixel-level annotations are available in the source modality, the self-supervised model outperforms the semi-supervised model. This implies that a larger and more optimal segmentation network, relying solely on supervision from annotated synthetic pseudo-target data generated by our modality translation network, yields better performance than a smaller model provided with additional weak labels. This observation raises that there is a trade-off between training a bigger model and training a semi-supervised model. However, when annotated source data is scarce (less than $50\%$ of annotated samples), the semi-supervised variant enables to preserve consistent segmentations and outperforms the self-supervised variant, even with a smaller segmentation network. Indeed, in this scenario training the model on actual target images becomes crucial as it has access to fewer synthetic samples and the generated pseudo-target images may be less reliable. Therefore we claim that MoDATTS has the potential to alleviate the annotation burden in cross-modality segmentation tasks.

Although producing the diseased and healthy image-level labels for 3D images does not add substantial annotation cost, it still represents a limitation when compared to unsupervised domain adaptation methods that rely only on pixel-level annotations available in the source modality. We thus specify that the self-supervised variant of MoDATTS does not require these weak labels to be trained, which is an advantage over the semi-supervised variant. In the end, the choice of either variant depends on the proportion of unannotated samples that the concerned dataset holds and the computational resources available.
\vspace{-0.2cm}
\paragraph{\textbf{Limitations}} As the pathologies we studied in this article (brain tumor and VS) mostly showed lesions on one side of the volumetric images, we were able to yield healthy and diseased samples by splitting the data into hemispheres. In real scenarios, full volumes showing healthy conditions should be collected to train the model. However, even though we experimented outside this setting we believe our results are representative of the actual behavior of the different models.\vspace{-0.2cm}
\\\\\
MoDATTS has demonstrated encouraging performance in the challenging task of cross-modality domain adaptation. However, it remains a heavy 3D method that requires training two distinct models (modality translation and segmentation). This represents high training times and consequent computational resources, specifically for the semi-supervised variant that contains several decoders and discriminators. Furthermore, unlike source-free models that solely rely on a segmentation model trained on the source modality for target adaptation, MoDATTS requires the presence of source images during training. However, it is worth noting that source-free methods under-perform in comparison to generative methods like MoDATTS and sometimes rely on image-level labels incurring substantial annotation costs \citep{source_free_3}.

\vspace{-0.2cm}
\paragraph{\textbf{Extensions}} We have tested MoDATTS on CrossMoDA and a modified version of BraTS, which both offer an ideal environment to test any domain adaption strategy for cross-modality segmentation. However they remain limited to segmentation between different MR contrasts. Further work will explore MR to CT adaptation. We even consider evaluating MoDATTS on a cross-pathology and cross-modality task. Specifically, we consider that leveraging annotated gliomas in FLAIR sequences from BraTS to segment intraparenchymal hemorrhages on CT scans \citep{mr_ct} is in the range of applications of our model. We believe that the semi-supervised variant of MoDATTS might provide benefits to mitigate the shift between these two conditions.

\section{Conclusion}

MoDATTS is a new 3D transformer-based domain adaptation framework to handle unpaired cross-modality medical image segmentation when target modality lacks annotated samples. We propose a self-supervised variant relying on the supervision from generated pseudo-target images and self-training, bridging the performance gap related to domain shifts in cross-modality tumor segmentation and outperforming other baselines in such scenarios. We offer as well a semi-supervised variant that additionally leverages diseased and healthy weak labels to extend the training to unannotated target images. We show that this annotation-efficient setup helps to maintain consistent performance on the target modality even when source pixel-level annotations are scarce. MoDATTS's ability to achieve 99\% and 100\% of a target supervised model performance when respectively 20\% and 50\% of the target data is annotated further emphasizes that our approach can help to mitigate the lack of annotations. The evaluation of MR to CT adaptation tasks will provide further insights into the potential applications of our approach.

\section*{Acknowledgments}
This research has been funded by the Natural Sciences and Engineering Research Council of Canada (NSERC), and the Canada Research Chair. We thank Compute Canada for providing the essential computational resources to complete this study. 

\bibliographystyle{model2-names.bst}\biboptions{authoryear}
\bibliography{refs}

\end{document}